# Physics and Chemistry from Parsimonious Representations: Image Analysis *via* Invariant Variational Autoencoders

Mani Valleti,[1,a] Yongtao Liu,[2] and Sergei V. Kalinin,[3,b]

[1] Bredesen Center for Interdisciplinary Research, University of Tennessee, Knoxville, TN 37916 USA

[2] Center for Nanophase Materials Sciences, Oak Ridge National Laboratory, Oak Ridge, TN 37831 USA

[3] Department of Materials Science and Engineering, University of Tennessee, Knoxville, TN 37916 USA

Electron, optical, and scanning probe microscopy methods are generating ever increasing volume of image data containing information on atomic and mesoscale structures and functionalities. This necessitates the development of the machine learning methods for discovery of physical and chemical phenomena from the data, such as manifestations of symmetry breaking in electron and scanning tunneling microscopy images, variability of the nanoparticles. Variational autoencoders (VAEs) are emerging as a powerful paradigm for the unsupervised data analysis, allowing to disentangle the factors of variability and discover optimal parsimonious representation. Here, we summarize recent developments in VAEs, covering the basic principles and intuition behind the VAEs. The invariant VAEs are introduced as an approach to accommodate scale and translation invariances present in imaging data and separate known factors of variations from the ones to be discovered. We further describe the opportunities enabled by the control over VAE architecture, including conditional, semi-supervised, and joint VAEs. Several case studies of VAE applications for toy models and experimental data sets in Scanning Transmission Electron Microscopy are discussed, emphasizing the deep connection between VAE and basic physical principles. All the codes used here are available at https://github.com/saimani5/VAE-tutorials and this article can be used as an application guide when applying these to own data sets.

[a] svalleti@vols.utk.edu
[b] sergei2@utk.edu

# 1. Introduction

Since the invention of the optical microscope by Levenhoek,[1] microscopy has played progressively growing role in science. Modern optical microscopy is the mainstream characterization tool in biology,[2] materials sciences,[3-6] nanotechnology,[7] and spans desktop tools to bespoke confocal and light sheet imaging systems. Subsequently, scanning probe microscopy has literally opened the door to the nanoworld,[8] and enabled visualization of structures as well as electronic,[9-11] magnetic,[12,13] chemical,[14] and electromechanical[15] functionalities. Electron microscopy has become the mainstay of materials science,[16] condensed matter physics,[17] molecular biology, chemistry and catalysis.[18]

Common for all areas of electron, optical, and probe microscopy imaging is the progressive transition from the qualitative imaging when only the salient features of image are qualitatively explored to quantitative imaging. In optical imaging, this is exemplified by the multitude of the biological imaging modalities, metallographic imaging, and broad variety of hyperspectral imaging modes including microRaman and photoluminescence measurements. In electron and scanning probe microscopy, the examples include mapping order parameter fields,[19] learning the internal structures of grain boundaries,[20] and decoding complex structures and defect chemistries of 2D materials. For scanning probe microscopy, this includes techniques such as molecular unfolding spectroscopy[21,22] that yields quantitative mechanisms of single-molecule reactions and piezoresponse force microscopy[23] that quantifies ferroelectric polarization dynamics.

Over the last decade, the volumes of data acquired over the modern imaging tools have grown exponentially, and now represent large images, multimodal images comprising multiple channels, hyperspectral data sets, and dynamic images such as videos. This transition to the quantitative imaging was largely enabled by the introduction of the computational systems for storing and analysis of the data on the level of the individual tools, facilities, and inter-facilities networks. At this point, virtually all microscope systems are equipped with the local data storage and analysis capabilities. While this transition is almost complete now, it is important to note that this trend is relatively new and is largely confined to the last 20 years. The emergence of the Google Colabs and Jupyter notebooks over the past 5 years have enable simple code and analysis workflow sharing between the individual researchers. Finally, the rapid growth of the cloud capabilities such as Amazon AWS, Google Cloud, or Microsoft Azure now opens the pathway for integration of data flows between multiple facilities and cloud-based storage and analytics of the imaging data, where only the analysis results are visualized on the personal devices.

The availability of these capabilities and the data sets that contain information about local structures functionality, and dynamics, necessitate the development of the data analysis tools for extracting the salient features of system behavior. Traditionally, such analyses have been based on the classical computer vision algorithms including local and semi local filters, Fourier, Radon, and Hough transforms,[24] *etc.* requiring extensive manual development and often elegant, but highly personalized, workflows. The emergence of the supervised deep learning methods in images[25] has offered an alternative approach and rendered many traditionally complex problems such as semantic segmentation, instance segmentation, *etc.* straightforward. This was done simply by translating workflows from computer and medical imaging to the microscopy domains. The advancements in the ensemble methods[26-30] have also allowed to address the out of distribution drift effects and opened the pathways to engender these methods as a part of real time microscope operation.[29,30]

However, the supervised methods, by definition, apply only if the physical picture of the phenomena of interest is known. Practically, in many applications scientists seek to discover new physical phenomena such as manifestations of symmetry breaking in electron and scanning tunneling microscopy images, variability of the nanoparticles, *etc.* These problems constitute the unsupervised physical discovery problems and necessitate the development of necessary algorithms. These in turn should satisfy the practical considerations present in the analysis of imaging data, *i.e.*, address the imaging system distortion, rotational invariances in image plane, allow for the fully unsupervised and semi-supervised learning tasks, and allow for combined clustering and elucidation of factors of variance.

Here, we review the principles and applications of the Variational Autoencoders (VAEs), a powerful method for the analysis of high-dimensional data. We discuss the basic principles and intuition behind VAE using the simple toy data sets and introduce several real-world examples in supervised, semi-supervised and unsupervised settings. We specifically emphasize the links between VAEs and physics of the system. We further discuss invariant autoencoders, both to discover the salient aspects of materials structure and allow for the distortions inevitable in real imaging systems. Finally, we discuss the VAE architectures including conditional, joint, and semi-supervised VAEs, and discuss the types of the image analysis tasks they can be applied to. All these methods are incorporated as a part of "VAE-tutorials" GitHub repository and is available for public access at https://github.com/saimani5/VAE-tutorials.

## 2. Variational autoencoders

Variational autoencoders have been introduced by Kingma[31,32] and Rezende[33] almost simultaneously in 2014 and are a special class of generative probabilistic models. They possess multiple unique properties that make them singularly useful for image analysis, and especially physics-based analytics. Below, we discuss the basic principles of VAEs and compare then to classical autoencoders, briefly introduce mathematical formalism beyond VAEs, and illustrate their behavior *via.*, simple examples. Here and throughout the article, we would like to highlight that, as for many machine learning (ML) tools, application of VAEs to real world problems requires development of both general and domain-specific intuition (how would they work for YOUR images), and hence we encourage the readers to experiment with the associated notebooks and use them for their own data.

### 2.a. Introduction to autoencoders

Prior to discussing the VAEs, we briefly introduce the principle of simple autoencoders (AEs) and their applications. Generally, AE represents a class of neural networks comprised of an encoder and a decoder part. The encoder takes the input object, such as an image or image patch, and passes it through a set of convolutional or dense layers with progressively decreasing dimensionality down to the bottleneck or latent layer. In this process, the dimensionality of the data is reduced from the original size to a usually very small number of latent variables. The decoder takes the latent vector and passes it through a set of layers of increasing dimensionality to reconstruct the original data. Often but not necessarily the architecture of the encoder and decoder are symmetric with respect to the bottleneck layer. The training process for autoencoders aims to minimize the difference between the input and output data, which helps the autoencoder learn to effectively compress and reconstruct the original data set. The loss function in the AE can be represented in equation 1, where *X* is the input image and *decoder(encoder(X))* is the reconstructed image and *LF* is the loss function.

$$Loss = LF\left(X, decoder(encoder(X))\right) \quad (1)$$

In principle, the AE can be trained with many loss functions, including but not limited to cross entropy loss, mean absolute error, and mean squared error. The structure of the encoder and decoder can follow any of the commonly used machine learning paradigms, such as fully connected multilayer perceptrons,[34] convolutional networks,[35] attention layers,[36] and graph networks.[37] The choice of the architecture is determined by factors such as the type of input data (*e.g.*, spectra, images, or graph representations) and the requirements of the specific domain. Similarly, the structure and architecture of the encoder and decoder can be different. However, the dimensionality of the input and output data are kept the same – alternatively, we have encoder-decoder architecture rather than autoencoder where output is different from the input (*e.g.*, images to spectra[38]). These architectures also have multiple applications in establishing structure-property relationships, quantitative structure- activity relationships, *etc*. but are outside the scope of present review.

At the first glance, the autoencoders seem to be a very counterintuitive application – after all, they try to reconstruct the data itself. However, the real power of the autoencoders is their capability to encode high dimensional data in the form of low-dimensional vector and decode the data from the latent vector. In other words, it builds the relationship (input) → ($l_1$, ..., $l_n$) and ($l_1$, ..., $l_n$) → (output), where input and output are ideally as close as possible.

This transformation allows for several useful properties and applications. First, encoding and decoding the data *via.* latent vectors select the relevant features in the data and reject the spurious elements. As a toy dataset to demonstrate this, we have chosen the MNIST dataset.[39] The MNIST dataset is a collection of images of handwritten digits between [0, 9] and is usually the go-to dataset for testing the computer vision algorithms. The results of the application of AE on the MNIST dataset are shown in Figure 1. The encodings of the original input in a low dimensional latent space can be visualized in two different ways. Firstly, each input image can be plotted using a scatter-point in the low dimensional dataspace using their latent vector. In this method, all the input images can be represented in a single scatter plot based on their corresponding location in the latent space. Additionally, we can color the points using their class label and see how different classes are clustered in the latent space. We will refer to such plots as the latent space distributions from hereon.

In the case of MNIST dataset, the latent space dimensionality is set to 2 and its corresponding latent space distribution is shown in Figure 1a, where the scatter-points are colored based on their class [0-9]. The latent space distribution shows that, even without any prior information about the class labels, the images belonging to one class are clustered in the latent space. Furthermore, the classes that share some common features overlap with each other.

The second approach to visualize the construction of latent space is shown in Figure 1b. In this approach, the input dataset is discarded after the training phase. The trained latent space is then uniformly sampled, and then decoded back into the input space. These decoded images are then plotted in their corresponding positions in the latent space. This way of visualizing the latent space will be referred to as the decoded latent space, or latent representation, and is shown in Figure 1b for the MNIST dataset. Decoded latent space helps us to visualize the characteristics of the latent space without using the input dataset.

For the MNIST dataset, the decoded latent space shows how the image classes are distributed in the latent space and the overlapping of features from different image classes. In AE, the latent space is not regularized, and the limits used in each of the latent dimensions to uniformly sample the latent space are obtained from the latent distributions. For the MNIST case, the limits for the decoded latent space are obtained for the latent distribution in Figure 1a and are set to [-100, 100] for both latent dimensions. The axis labels on the decoded latent are omitted in certain places to make the figures legible. The purpose these decoded latent spaces serve is to show the variation of the inputs in the latent space. Additionally, since the latent space is uniformly sampled, the absolute values of the latent variables do not contribute to additional understandings.

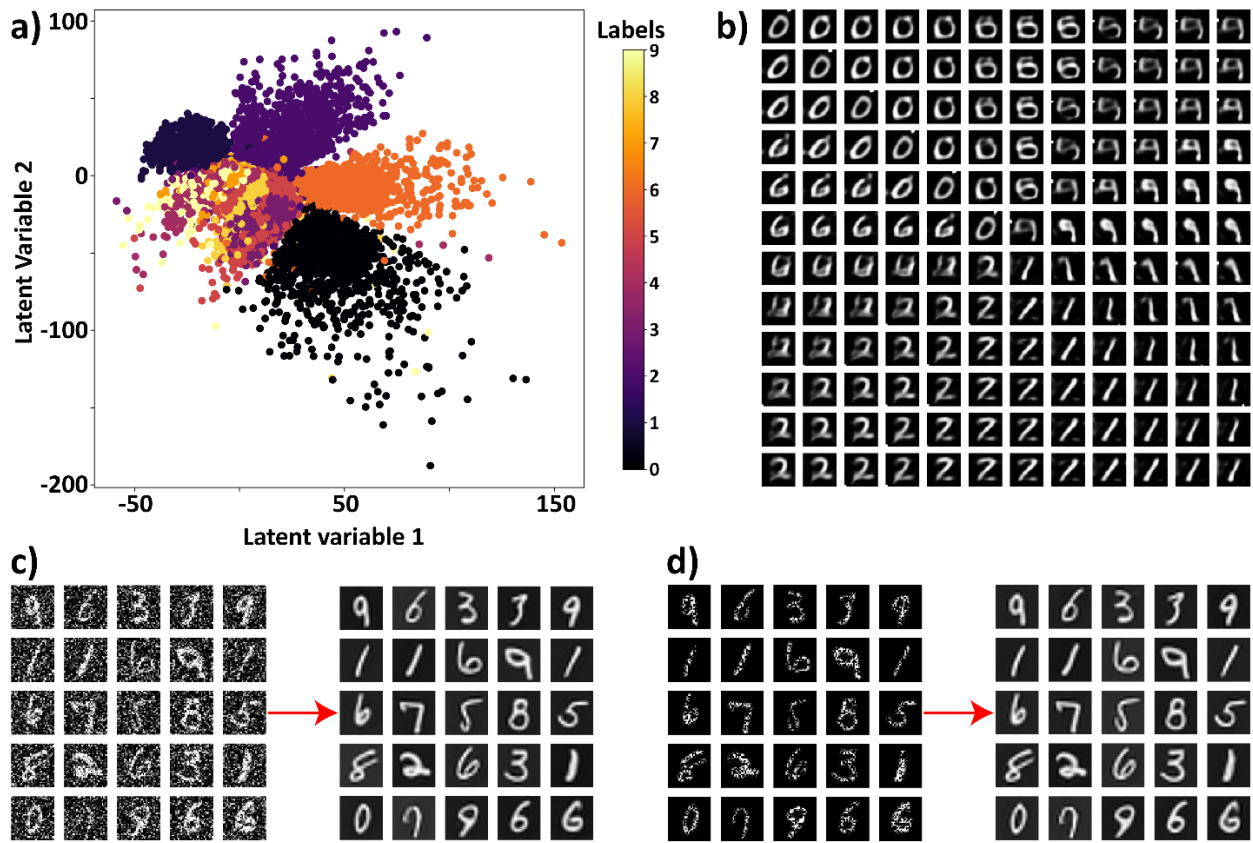

**Figure 1.** (a) MNIST dataset represented in the 2D latent space and colored using the class labels, (b) uniformly sampled and decoded latent space, (c) representation of denoising the MNIST dataset using AE, the noisy inputs to the AE and corresponding cleaned images are shown, (d) representation of filling the missing pixels application on the MNIST dataset using AE, the inputs with missing pixels and the reconstructed outputs from the AE are shown.

The AEs can be extensively used for denoising and data reconstruction. For this and many other applications, the key element becomes the choice of the inputs and outputs. For example, for denoising, a noisy dataset is used as the input and the loss function in equation 1 is evaluated between the reconstructed dataset (output) and a clean dataset. This method works if we have access to the noise-free (clean) dataset beforehand. One way to overcome this limitation is to create a synthetic dataset using the domain expertise so that it closely represents the dataset at hand. One can then add noise to the dataset and train the autoencoder on the synthetic dataset. This trained network can then be used to clean the original dataset. We demonstrated this by applying noise to the MNIST dataset and used the AE architecture to clean the noisy images. The inputs (noisy) and outputs (clean) for this process are shown in Figure 1c.

Another application of autoencoders is image reconstruction, *i.e.*, inpainting missing pixels or restoring color. In the missing pixels case, the input to the autoencoder is the dataset with incomplete data and the output is the cleaned and complete dataset. The missing pixels input of the MNIST dataset, and its corresponding filled-output are shown in Figure 1d. The analyses that constructed the Figure 1 are available with code in "Intro to AE and VAE.ipynb" Jupyter notebook that accompanies the manuscript. Each section has once such notebook available with it where the dataset used in the analysis can be downloaded into the notebook. The reader can reproduce the entire analysis corresponding to the results discussed in the manuscript. The weights of the architecture that are either time or memory extensive are loaded in the manuscript to accelerate the process.

It is also very important to note that this process implicitly illustrates the limitations of autoencoders – the missing information is not coming from nowhere – it in fact is derived from the correlations between the features in the original data set. If the objects with the shape of the cars were red in the training data set, then recolored cars will also be red in the reconstructed data set. However, if the data set was acquired during the Henry Ford times ("you can buy the car of any color, as long as it is black"), the reconstruction will be misleading. This is one of very significant limitations of the neural networks applied to reconstruction and super-resolution problems, and these and other limitations will be repeatedly pointed throughout the text.

**2.b. Introduction to VAEs**

With the above-mentioned elucidation on AEs, we now discuss the VAEs, the main topic of this review. The encoder and the decoder part of VAE look similar to those of autoencoders, with a significant difference being the structure of the latent layer. In this case rather than rigidly pass the data through the latent bottleneck, the VAE samples the latent space. The excellent introduction into principles and properties of VAEs are available in the recent publication by Kingma.[32]

VAEs have the same properties as autoencoders AEs, but also have the ability to disentangle the latent representations of data. While the formal definition of this process is still debated, it is generally understood to involve the association of different factors of variation in data with different latent variables. Variational Autoencoders (VAEs) have been widely recognized for their generative capabilities, where they can generate new, unseen data samples that are similar in distribution to the training data, making them a powerful tool in the realm of unsupervised learning. By sampling a latent point that lies between the latent representations of two input data samples, VAEs can produce new data samples that have features from both input data points, but that are not part of the input dataset. These newly generated samples can be thought of as a smooth interpolation between the input data samples in the latent space, and they can provide valuable insights into the structure of the input data distribution.

The MNIST dataset used in elucidating AEs is also used here to explain the capabilities and shortcomings of the traditional VAE construction. We use the decoded latent space and latent distributions as the visualization tools for analyzing the results of VAE. Decoded latent space and latent distribution of the VAE trained on the MNIST dataset are shown in Figures 2a and 2e, respectively. The limits for plotting the decoded latent space of the VAE and its variants for the rest of the manuscript will be [-1.5, 1.5] unless otherwise specified. These limits work because of the prior distribution enforced on the latent variables. The prior distribution is usually considered to be independent zero-centered normal distribution with unit variance which will be further discussed in the formalism section. The decoded latent space of the VAE shows how the important features of the dataset are encoded in the latent space. For example, all the straight lines segments involved in the dataset ("1"s and "7"s) are towards the left of the dataset which slowly transform to the circles towards the right of the dataset ("0"s, "6"s, and "8"s). The latent distribution shown in Figure 2e shows how the latent space is regularized, *i.e.,* different data classes are tightly packed in the latent space, with very narrow boundaries of "unphysical" structures between the regions corresponding to different digits. This paves way for the generative capabilities of VAE.

The shortcoming of the traditional construction of VAEs can be illustrated by adding additional variability (rotated MNIST) to the dataset. Generally, we choose this special case since in imaging (unlike the human handwriting) the orientation of the object in the image plane is apriori unknown. In the rotated MNIST dataset, each image is rotated by an angle that is randomly sampled from the range [-90°, 90°]. The resulting dataset is shown in Figure 2c which acts as an input to a different VAE network. The decoded latent space and the latent distributions of the VAE trained on this rotated dataset are shown in Figure 2c and Figure 2f respectively. The decoded latent space (Figure 2c) encodes features at varying angles, as well as the specific features discussed in the unrotated dataset case. The latent distributions in Figure 2f are colored by the class labels of the dataset. The class labels in the latent space are now ambiguous due to rotational symmetries present in some of the classes. The latent distribution shows that the VAE does a sub-optimal job in segregating the classes as no clear clusters for each class are visible. Increasing the number of latent variables in this analysis might help but there is no assurance that the latent space disentangles the angle involved in producing the image.

To counter this problem, we used a rotationally invariant variational autoencoder (rVAE) to analyze the dataset. This class of the variational autoencoders ensures that one of the latent variables is the angle of rotation of the object in the image. This angle-invariance (as well as translation-invariance) will be discussed in more detail in the section on rVAE. The decoded latent space and the latent distributions of the rVAE are shown in Figure 2d and Figure 2g respectively. The results of rVAE on the rotated MNIST dataset are shown here as we will use additional datasets to explain the workings of different classes of variational autoencoders after this section. The decoded latent space of the rVAE (Figure 2d) shows that all the digits are encoded with respect to the horizontal axis in a single orientation. The horizontal orientation in the latent is because the zero of the encoded angles is arbitrary in the rVAE. The latent distributions (Figure 2g) show clusters of different classes in the latent space distribution which shows that the rVAE does a better job at separating different classes in comparison to vanilla VAE. The analyses that constructed the Figure 2 are available with code in "Intro to AE and VAE.ipynb" Jupyter notebook that accompanies the manuscript.

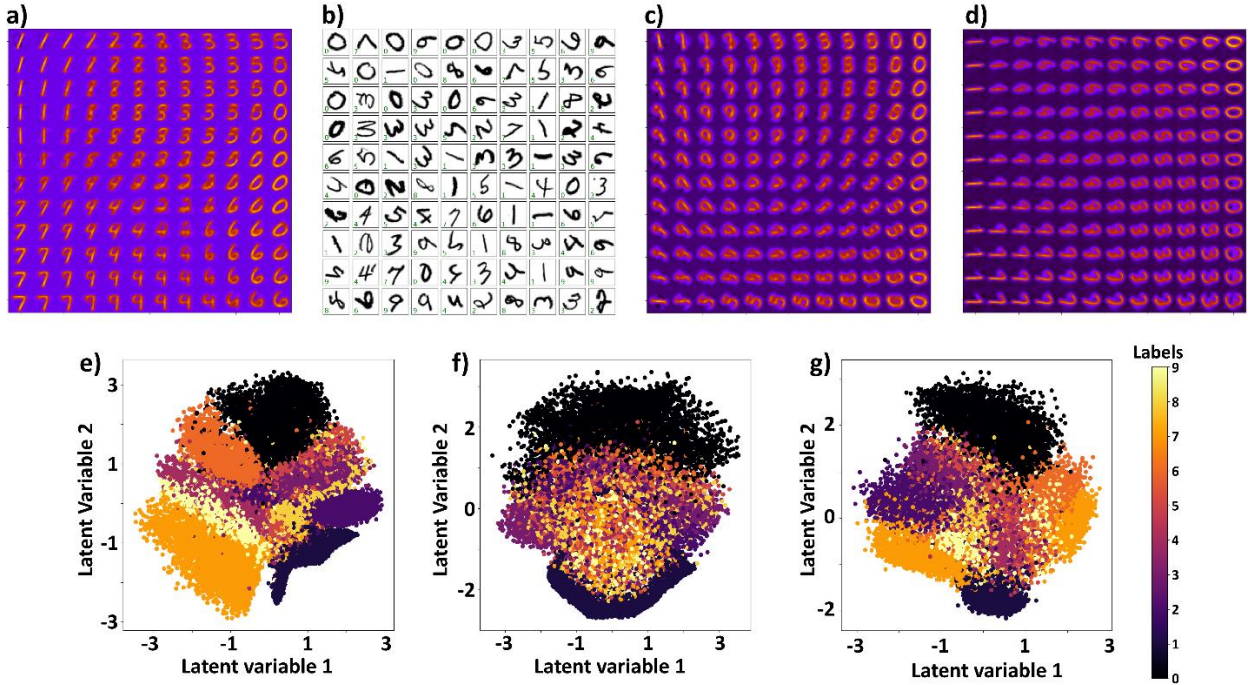

**Figure 2.** (a) The decoded latent space of the VAE trained on the MNIST dataset, the limits on each latent dimension are [-1.5, 1.5] for all VAE decoded latent spaces unless otherwise specified, (b) Examples from the rotated MNIST dataset where the angle of rotation of each image is randomly sampled from the interval [-90°, 90°], (c) Decoded latent space of the VAE when trained on the rotated MNIST dataset, (d) Decoded latent space of rVAE when trained on the rotated MNIST dataset. The latent distributions colored using the class labels of the images for (e) VAE trained on MNIST dataset, (f) VAE trained on rotated MNIST dataset, and (g) rVAE trained on the rotated MNIST dataset. The color bar for (e), (f), and (g) is the same and is shown in (g).

## 2.c. Mathematical formulations of VAE

The Variational Autoencoder (VAE) is a generative model that learns the latent space of high-dimensional data in an unsupervised way. The VAE consists of two parts: an encoder $q_\tau(z|x)$ also known as the inference network that is parameterized by $\tau$, and a decoder $p_\phi(x|z)$ also known as the generator network parameterized by $\phi$. The posterior distribution $p(z|x)$ is approximated by a tractable distribution which is often referred to as the variational distribution. For the analyses in this article, the variational inference is a Gaussian distribution whose parameters are predicted by the encoder $q_\tau(x|z)$. The decoder when provided with a latent space vector ($z$), outputs the parameters to the likelihood distribution $p(x|z)$. The goal of the VAE is to learn a compact and smooth latent space that can be used for generating new data samples.

To achieve this, the VAE introduces a prior distribution $p(z)$ over the latent space, typically a standard Gaussian distribution $\mathcal{N}(0, I)$ with zero mean and unit variance. This prior distribution acts as a regularizer, ensuring that the latent space is smooth and continuous. The encoder network maps the input data to a mean and a standard deviation of a variational Gaussian distribution in the latent space that approximates the latent space. The mean and standard deviation are used to sample from the Gaussian distribution in the latent space, allowing the VAE to generate new samples by sampling from this distribution and decoding them back into the original data space.

To train the VAE, the Evidence Lower Bound (ELBO) is maximized. The ELBO for VAE consists of two terms: the reconstruction loss and the KL divergence between the learned latent distribution and the prior distribution. The reconstruction loss measures the difference between the input data and the reconstructed output, while the KL divergence encourages the learned latent distribution to be close to the prior distribution, ensuring that the latent space is smooth and continuous. The VAE learns the optimal encoder and decoder networks by maximizing the ELBO given by equation 2.

$$ELBO_{vae} = -Reconstruction\ error - D_{KL}(q(z|x)||\mathcal{N}(0, I)) \quad (2)$$

In summary, the VAE is a generative model that learns a smooth and continuous latent space by introducing a prior distribution over the latent space and maximizing the ELBO. The encoder maps the input data to a mean and a standard deviation of a Gaussian distribution in the latent space, and the decoder maps the latent space to the output data. The VAE is trained by minimizing the reconstruction loss and the KL divergence between the learned latent distribution and the prior distribution or by maximizing the ELBO.

**2.d. VAE with the cards**

With the discussions of the mathematical foundations, we now proceed to test the capabilities of VAEs on a toy dataset that will be used to evaluate different versions of VAEs from here on. The toy dataset we have chosen for this purpose is of playing cards, with monochrome images of all 4 suits *viz.*, clubs, spades, diamonds, and hearts. Despite its simplicity, the cards dataset forms an interesting collection of symbols. For example, spades and clubs (without tail) have three-fold and mirror symmetry. Moreover, horizontally flipped hearts and spades only differ by small details which makes it hard for the unsupervised algorithms to differentiate them. Diamonds, on the other hand have a shape that is distinct from other three suites. Rotating the diamonds suit by $90^0$ is equivalent to uniaxial compression and resizing. This phenomenon introduces degeneracies in the outcomes of the application of affine transformations on this dataset. With these profound complexities in terms of symmetries, the cards dataset poses to be a challenging one for the machine learning algorithms – but at the same time the relevant behaviors can be readily illustrated in the 2D latent spaces, allowing for intuitively clear explanations.

The variability in the images of the card dataset is introduced using different affine transformations on each image. Each image is rotated using an angle that is randomly sampled from the range [-12°, 12°] for the datasets with low rotations and [-120°, 120°] for the datasets with high rotations. The images are additionally sheared equally in both x and y directions where the value of the shear is sampled from [-1°, 1°] for the low shear case and [-20°, 20°] for the high shear case. In total, four datasets are formed with combinations of these two transformations. The datasets thus formed are i) low rotations and low shear, ii) low rotations and high shear, iii) high rotations and low shear, and iv) high rotations and low shear. These datasets will be used to analyze different types of VAEs in the upcoming sections and will be referred to as cards dataset (i-iv) for brevity.

One network for dataset is trained and the corresponding results of all four datasets are shown in Figure 3. The figure is divided into 4 parts *viz.*, (p-s) corresponding to cards datasets (i-iv). Twenty-five randomly chosen examples from each dataset are shown in Figure 3a. The decoded latent space of all the four datasets is shown in Figure 3b. While the latent distributions of all four datasets are shown in Figure 3c where the scatter points are colored using the original class label of each datapoint. The limits for *x* and *y* axes in the latent distributions are omitted for brevity since the plots are only used for visualizing the clustering in the latent space. It can be observed from the latent distributions that the VAE does an optimal job in encoding the images into the latent space with low rotations (Figure 3cp and Figure 3cq). Clusters of four different classes in the dataset and the boundaries between them are clearly visible. The same conclusion can be drawn from the decoded latent spaces (Figure 3bp and Figure 3bq) of these two datasets.

Additionally, since the latent space in case VAEs is smooth, each cluster is expected to have a smooth variation in angles and shear in the latent space. In the decoded latent spaces, each image unambiguously belongs to one of the four classes which also aids us in making the conclusion that the VAE can generate optimal encodings for all the classes in the latent space for the first two datasets.

This is not the case for the latter two datasets (cards – iii and iv) and the results of these two datasets are shown in Figure 3r and 3s respectively. The latent space distributions ((Figure 3cr and Figure 3cs) show that only the diamonds class form a clear cluster and there is a considerable overlap between the other three classes. This overlap arises from the different symmetries present in the input dataset at high rotations. The diamonds do not have a significant overlap with the other three classes in terms of their structure. This can also be observed in the decoded latent spaces (Figure 3br and Figure 3bs) where we have images that have characteristics from two or more classes except for diamonds which occupy the top rows in both the cases. This analysis elucidates the short comings of VAE on the datasets with rotations especially in the presence of rotational symmetries. The complete analysis can be visualized and reproduced using the 'intro to AE and VAE.ipnyb' Jupyter notebook that accompanies this article.

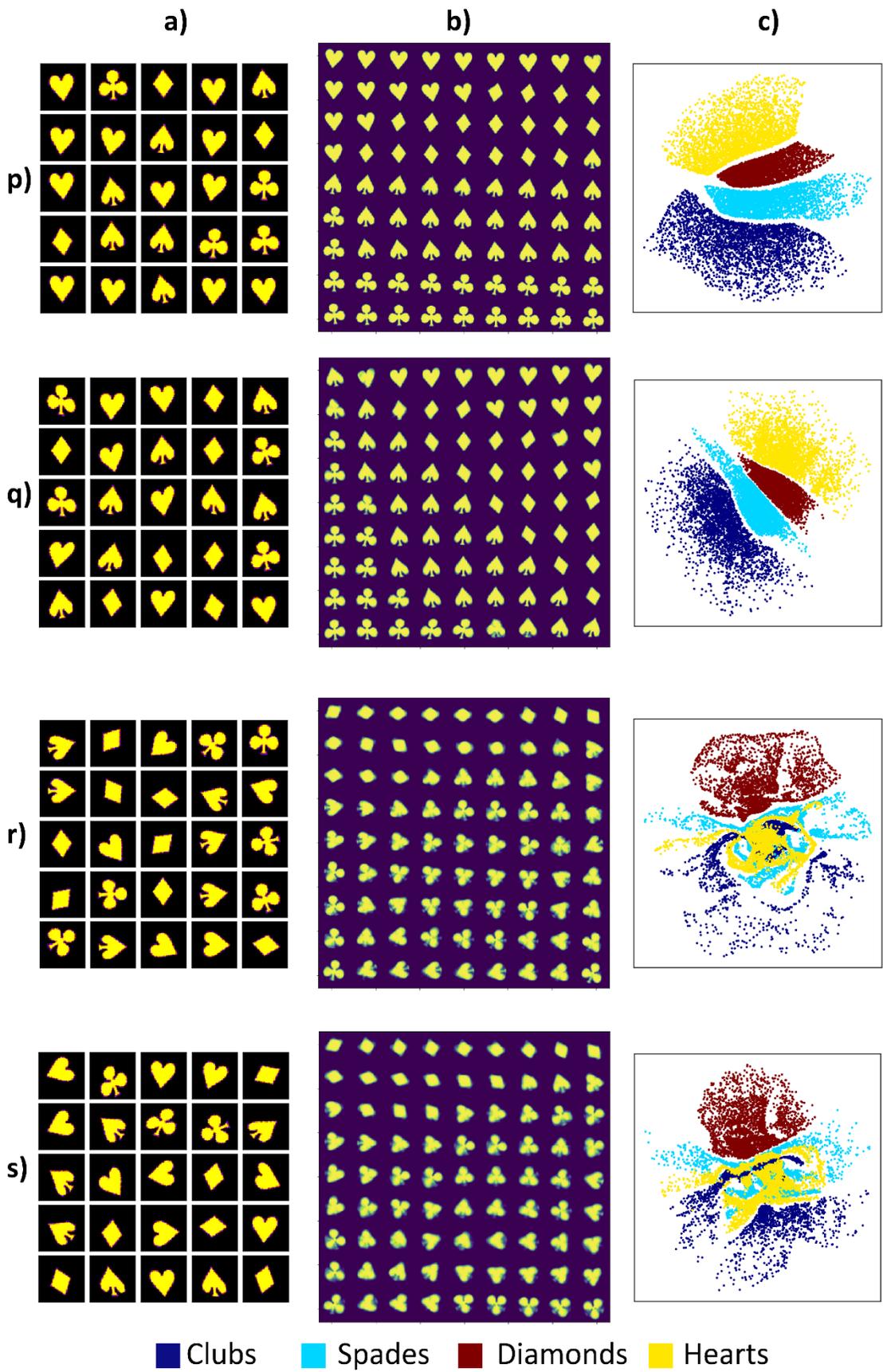

**Figure 3.** The rows in the figure *i.e.*, (p-s) correspond to the different cards datasets (i-iv) formed with the application of transformations on the images of the cards respectively. (a) Twenty-five randomly chosen images from each dataset to visualize the intricacies in each dataset, (b) decoded latent spaces of each dataset, and (c) latent distributions in the latent space for each dataset where the scatter points are colored using the original class labels.

## 3. Choosing the descriptors

The key element in application of VAE for imaging data is the choice of descriptors. For spectral imaging data $R(x, y, t)$, where $x$ and $y$ are coordinates. The natural object is the spectral dataset $R(t)$ at location $[x, y]$, where the $t$ is a descriptor that can be a physical parameter extracted from the spectra. For imaging data, there are multiple strategies for selecting the descriptors. One of them is to hand-pick the physical quantitative descriptors from the dataset based on the domain expertise. To elucidate this process, we consider the Piezoresponse Force Microscopy (PFM) image data showing ferroelectric domains of $PbTiO_3$ (PTO) at two different times $t_0$ and $t_1$ in Figures 4a-d. The amplitude responses of the PFM for both times are shown in Figure 4a and 4c respectively. The respective domain walls are semantically segmented using a deep convolutional neural network and are shown in Figures 4b and 4d respectively. In this case, the physical descriptors can be (1) the polarization state in a region, *e.g.*, the ratio of the bright region $P_1$ and the dark region $P_2$ in Figure 4a and 4c; (2) number of domain walls, *e.g.*, there are two domain wall $W_1$ and $W_2$ in the images Figure 4b and 4d; (3) the distance of two nearest domain walls, *e.g.*, $d_1$ and $d_2$ in Figure 4b and 4d. A different way of selecting the quantitative information from the image dataset is to use the neighborhood of the atoms in the atomically resolved STEM image. However, the efficiency of this method depends on presenting the entire information of the neighborhood in a format that can be used by the ML techniques. Molecular graphs,[40] SMILES,[41] SELFIES[42] are some of the well-known techniques to encode the neighborhood information in a format that can be handled by ML/DL techniques. These methods of selecting hand-picked descriptors or encoding neighborhoods oftentimes fail to include the entire information present in the dataset and/or result in unintended biases.

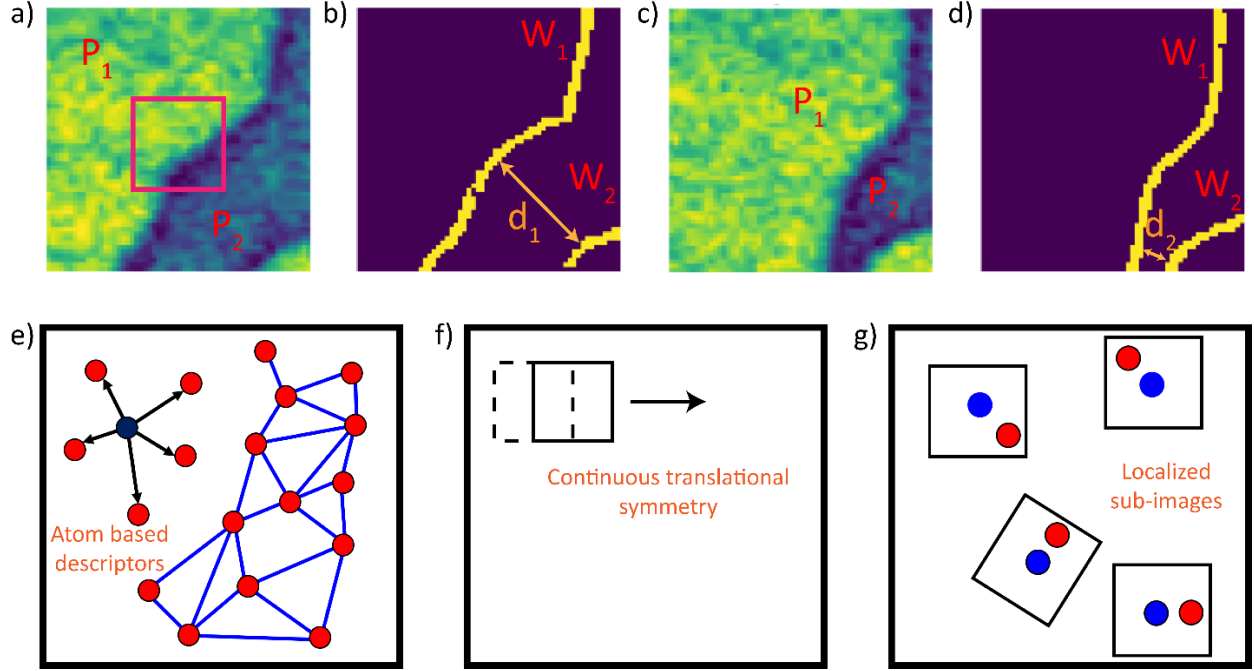

**Figure 4.** (a-d) Piezoresponse Force Microscopy image of ferroelectric PTO film, adapted from ref.[43] PFM amplitude image at time (a) $t_0$ and (c) $t_1$, domain wall images generated by DCNN corresponding to time (b) $t_0$ and (d) $t_1$. Schematic of choosing the neighborhoods of atoms in atomically-resolved images as descriptors. The feature engineering in the VAE image analysis includes selection of image patches. These can be based on rectangular sampling grid (f) (sliding window approach) or be centered on individual features such as atoms (g) or domain walls (red square in a). For dynamic data, the feature can be centered on the time-delayed image. Finally, the patches can be analyzed with or without rotational, translational, or other symmetries depending on the physics of the physical problem and imaging process.

The simpler way of selecting the descriptors of the system is the construction of the sub-images centered on the square grid within the original image. The density of the square grid determines the number of sub-images sampled from a given image. The schematic of this method is shown in Figure 4f. This method precludes any user-specific biases during the construction of the descriptors. The construction of these sub-images is straightforward. This approach has been used in several previous studies[44-46] However, for images containing object of interest the relationship between the patch center and the object can be arbitrary. Correspondingly, these random shifts become the primary factors of variation within the data. For example, Figure 4a-d shows a PFM data of ferroelectric PTO film, during the PFM measurement, a DC bias is applied through the tip to switch the polarization. The polarization state variation and domain wall motion can be recorded by continuous PFM image scan. Figure 4a shows a frame at time $t_0$ of this continuous PFM dataset. Sub-images can be extracted from this dataset, an example corresponding the red square marked region is shown in Figure 4a. Since the polarization state and domain wall are changing under the application of DC bias, after a certain time (*e.g.*, at time $t_1$), a polarization state change (Figure 4c) and a domain wall shift (Figure 4d) can be observed. This can be partially offset by the shift-invariant VAE[47] approach as will be discussed below. However, shift-VAE is limited when the object shape changes due to overlap with the patch edge, *etc*.

The second approach for construction of sub-images combines the patch generation with the physics-based descriptions. For STEM data, it is natural to define patches centered at the individual atoms (Figure 4g). In this case, the patch collection contains the full information in the image, but *via.*, construction analyzes the neighborhoods of individual atoms. This in turn necessitates introduction of rotationally invariant autoencoders, since otherwise the objects can have arbitrary orientation in the image plane.

Finally, for the time-dependent dynamic data the descriptors can be constructed to contain time information. For example, when analyzing the ferroelectric domain wall motion, the patches can be centered at the time-lapsed images. In this manner, the descriptors will contain the information on the time dynamics.[43,48]

## 4. Building invariant VAEs

The examples of the VAEs above illustrated the principle of VAE and the choice of descriptors in elucidating the phenomena of interest. However, applications of VAEs for image analysis is associated with the additional challenges. As discussed above, ultimately, we aim to discover the factors of variability in data or disentangle the data representations. However, in real images many factors of variability will be associated with the positions of the object of interest, rotations, or small changes of scale. For example, when identifying nanoparticle shapes from microscopy images, we generally assume that nanoparticle will be the same down to rotation in image plane. At the same time, uncertainties in establishing anchor points when choosing descriptors will result in the presence of translational offsets in images. These considerations prompt the development of invariant VAEs, generally referring to the network architectures that separate physical factors of variability (translation, rotation, scale) in separate latent variables, and disentangling the factors of variability in remaining degrees of freedom.

## 4.a. Introducing invariance

In this section, we explore the construction and workings of rotationally invariant autoencoders, which we address as rVAE from hereon for brevity. In the traditional setting of VAE, the latent variables encompass the essential features of the input dataset and are usually not controlled by the user. For example, in the MNIST dataset, the input datapoint's writing style and class (0-9) are the dominant features learned by the latent space. The rotations and translations of the object present in the images are treated as the dataset's features. The latent space learns them by treating them as input variations along with other features. For example, a single digit from the MNIST dataset can be rotated and translated randomly, and a VAE can be trained on this dataset. In this case, the angles and the translations are expected to be encoded in the smoothly varying latent space. In this process, the VAE treats the same object with different rotational and translational transformations as separate entities and produces different encodings for them. The rVAE network is designed such that 1-3 latent dimensions is used to explicitly encode the rotations and/or translations present in the input image. By incorporating both rotational and translational invariance in the rVAE network, the network can explicitly encode both rotational and translational information in the latent space, providing a more robust and accurate representation of the input data. For example, the encodings of images with similar objects appearing at different orientations will be similar in the latent space except for the encoding in the dimension that explicitly encodes the angles.

Rotations and translations are two important aspects of the images we encounter in our daily lives, as well as in many scientific applications. In fields such as materials science and condensed matter physics, the ability of the rVAE network to capture both rotational and translational information can be particularly useful for analyzing microscopic images. For example, by using rVAEs to analyze microscopic images, researchers can better understand the underlying rotational and translational patterns and their impact on material properties. Similarly, in condensed matter physics, rotations and translations play a crucial role in understanding the behavior of particles at the atomic and molecular scale, and rVAEs can be used to analyze and generate diverse images of these systems, providing insights into both their rotational and translational dynamics. Overall, the incorporation of both rotational and translational invariance in the rVAE network provides a powerful tool for analyzing and generating diverse images in various scientific applications, enabling researchers to better understand both the rotational and translational patterns and dynamics in their data.

The encoder part of the rVAE is similar to the traditional VAE, where a fully connected dense neural network (DNN) or a convolutional neural network (CNN) progressively decreases the dimensionality of the dataset until we reach a bottleneck layer of k-dimensions. We will refer to these latent dimensions of different variants of VAEs including rVAEs as $[z_1, z_2, …, z_k]$. These latent dimensions capture the variabilities in the input dataset and smoothly encode them in the latent space. The rVAE has the 1-3 additional features concatenated to this bottleneck layer corresponding to the angle ($\theta$), x-translation ($\Delta x$), and y-translation ($\Delta y$), depending on the type of rVAE used for training. Hence the dimensionality of the latent space of the rVAE is $k+3$ for implementing rotational and translational invariances, where $k$ is the number of latent variables chosen by the user. rVAEs use a special decoder known as the spatial generator net introduced by Bepler et al.[49] in 2019. The spatial generator decoder net uses the coordinates of the input image along with the low dimensional latent representation of the encoder to decode the images. Initially, we form a 2D array of the coordinates of the pixels involved in the input image. The latent features corresponding to the rotational and translational invariances ($\theta, \Delta x, \Delta y$) are then used to rotate and translate the 2D coordinate array, and this transformation is given by equation 3,

| | $$x_t = x * \begin{bmatrix} \cos\theta & \sin\theta \\ -\sin\theta & \cos\theta \end{bmatrix} + [\Delta x, \Delta y]$$ | (3) |

Where $x$ and $x_t$ are the original and transformed coordinates, respectively, $\theta, \Delta x, \Delta y$ are the rotational and translational encodings of the rVAE.

The first layer of the decoder projects the latent space encoding $[z_1, z_2, z_3 … z_k]$ into a higher dimensional space. The transformed coordinates are then projected into this higher dimensional space using a different neural network. After the first layer, the higher dimensional z-vector is made into $N$ copies, where $N$ is the number of coordinates or pixels in the input image, and each higher dimensional coordinate vector is added to the copies of the z-vector. After this operation, we have a higher dimensional encoding for each of the pixels. We then decode them using a fully connected DNN or a CNN like in the VAE, where the neurons/filters in the output layer are one. By making N copies of the z-vector, adding the higher dimensional coordinates to each of the copies, and decoding them back into the real space, the decodings can be viewed as conditional probabilities on each pixel where each decoder neuron is conditioned on the transformed coordinate. The schematic of rVAE is elucidated in the Figure 5a. The parameters for the encoder and the decoder for rVAE are learned by maximizing the ELBO given by equation 4.

$$ELBO_{rvae} = -Reconstruction\ error - D_{KL}\big(q(z|x)||\mathcal{N}(0,I)\big) \quad (4)$$
$$- D_{KL}\big(q(\theta|x)||M(0,\kappa)\big) - D_{KL}\big(q(\Delta x|x)||\mathcal{N}(0,\sigma_s)\big)$$

Where reconstruction error is the error between the reconstructed image and the original image which can be computed using the loss functions like mean squared error, cross entropy, and so on; the second term refers to the KL divergence between the posterior and the prior ($\mathcal{N}(0,1)$) of the latent representations; the third term refers to the KL divergence between the posterior distribution of the angle and the von-Mises prior on $\theta$ with parameter $\kappa$; and the last term in the KL divergence between the posterior of translations and its normally distributed prior with a standard deviation $\sigma_s$. The parameters involved in the prior distributions of these latent variables are hyperparameters of the network and the domain knowledge about the input dataset aids us in choosing them. The weights of the entire architecture are learned by performing gradient ascent on the *ELBO$_{rvae,}$* as discussed in the VAE section.

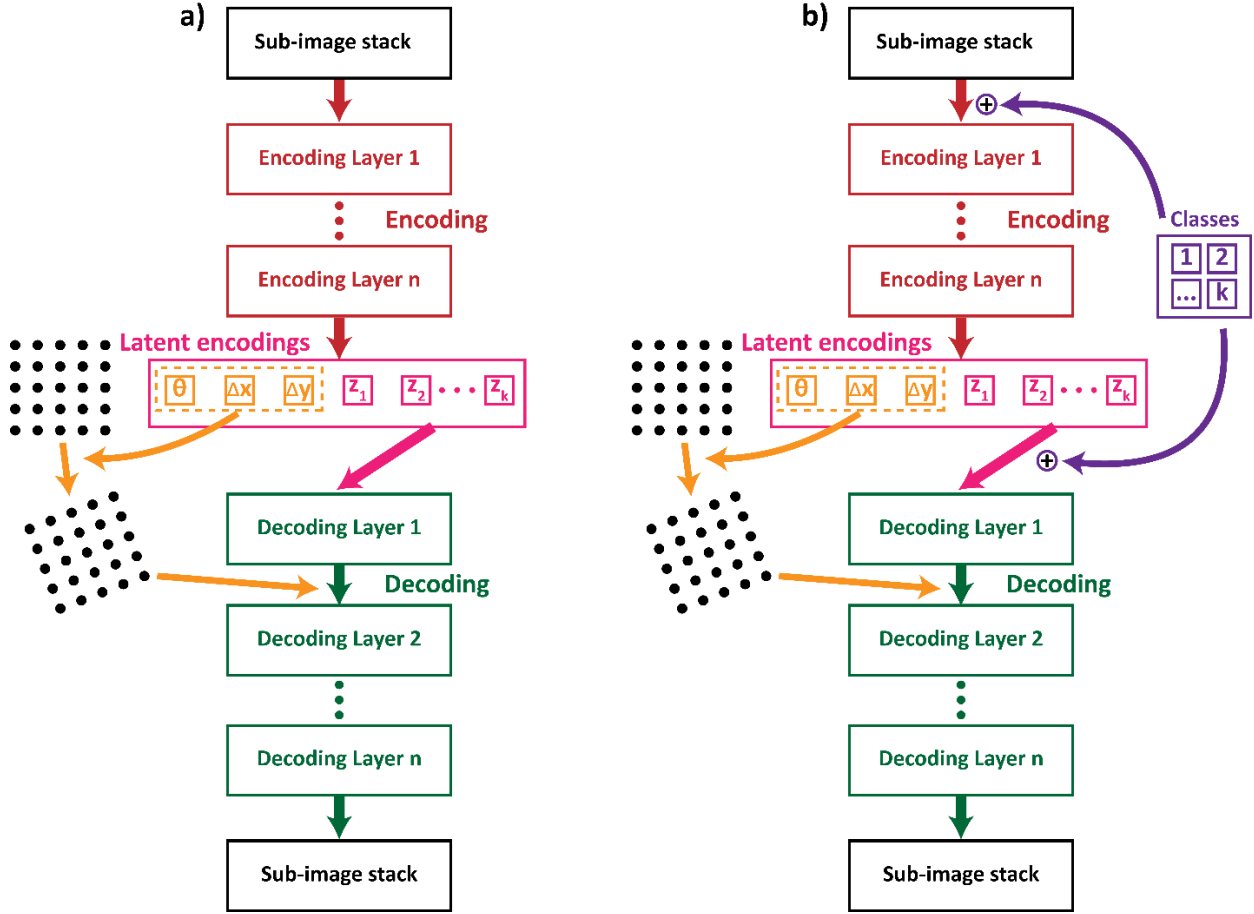

**Figure 5.** Working schematics of (a) rotationally invariant variational autoencoder (rVAE) and (b) conditional-rotationally invariant variational autoencoder (crVAE).

### 4.b. rVAE on cards dataset

To explore the application of rVAE, we use the cards datasets i-iv discussed earlier in the VAE section. Datasets are constructed from the original four cards belonging to the four suits, namely spades, diamonds, clubs, and hearts, by randomly rotating and shearing them. The random values for rotations and shears are picked from the interval [-$\alpha$, $\alpha$] and [-$s$, $s$], respectively. The specific values for $\alpha$ and $s$ for all four datasets are discussed in the VAE section. One rotationally invariant network per dataset is trained with 2 hidden layers in both the encoder and the decoder. The low shear cases *i.e.*, cards datasets i and iii use 256 neurons in the hidden layers and the high shear cases (cards dataset ii and iv) use 512 neurons in their hidden layers. The encoder-decoder architecture is symmetric throughout this article. The rVAEs are expected to explicitly disentangle the rotational features and form a smoothly varying latent space that encodes the other variations in the dataset, which is the shear angle in all four examples with varying limits. The first two cases with low rotations are not discussed in the manuscript as they are straightforward and can be found in the Jupyter notebook that accompanies the manuscript. Additionally, a simple VAE did an adequate job for these two datasets. The latent space forms a smooth variation of the cards, and the cards were all assigned to different areas of the latent space. The results for the last two cases with high shear are discussed in Figure 6.

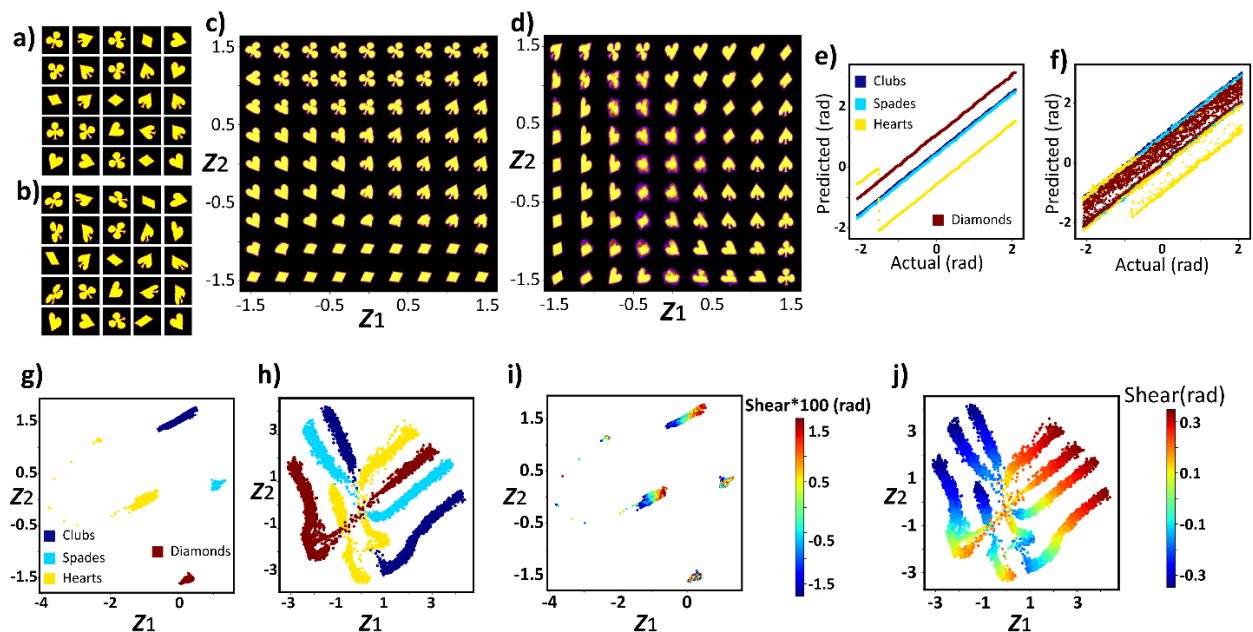

**Figure 6.** Twenty-five randomly sampled images form the (a) cards dataset iii and (b) cards dataset iv, decoded latent space correspond to the (c) cards dataset iii and (d) cards dataset iv, encoded angle *vs*. ground truth angle of the (e) cards dataset iii and (f) cards dataset iv where the points are colored using the class labels, latent distributions of the (g) cards dataset iii and (h) cards dataset iv where the points are colored using the class labels, variation of ground truth shear value in the latent space for all the four classes corresponding to (i) cards dataset iii and (j) cards dataset iv.

Figures 6a and 6b show the example images from both the high shear datasets with low rotation (cards dataset iii) and high rotations (cards dataset iv) applied to the input images. The decoded latent space for two different rVAEs trained is shown in Figure 6c and 6d, respectively. For the decoded latent spaces corresponding to the invariant VAE, the latent variables that capture the angles and translations are set to zero. This gives us the insight into how the images from each class are encoded in the latent space without the rotational and translational transformations. The baseline zero of the decoded angles is a function of the class label of the input image.

This can also be observed from the decoded latent spaces where each class has a different orientation in the latent space. The predicted angle of rotation *vs.* the actual angle involved in producing the input dataset are plotted in Figures 6c and 6d for the datasets iii and iv respectively. The color of the points corresponds to the class label of each input image. Finally, the latent distributions for the 2 cases are shown in Figures 6g-j. Here the latent distributions are colored using the original class labels in Figures 6g and 6h, using shear value in Figures 6i and 6j for the two datasets. The latent distributions show that the rVAE is able to distinguish the four different classes present in the low shear dataset (cards dataset iii) and formed clusters in the latent space.

The ambiguous images in the decoded latent space emerge because the given points reside midway between the points in the input dataset. The datapoints from different classes are not overlapped in the latent distributions plot. Only the hearts suit is identified as two different clusters. The reason for this can be seen in the decoded latent space (Figure 6c) towards the left ($z_1 = -1.5$), where the hearts suit is decoded in two different rotations. One of these decoded hearts suits has similarities with clubs towards the top, and the other has similarities with the spades and can be found in the middle of the decoded latent space. This can also be inferred from the angles plot (Figure 6e) where two lines with different slopes are allocated to the hearts suit. It can also be seen from this plot that the baseline zero of the encoded angle is different for different classes. Finally, Figure 6i shows that, for each class, the ground truth shear angle varies smoothly in the latent space. The deviation from this behavior is observed in cards dataset iii. This might be because the range of the applied shear is small compared to the angle of rotation for this dataset.

For the high rotation and the high shear dataset, the rVAE still outperforms the vanilla VAE by forming clusters in the latent space (Figure 6h). Although the clusters formed in the latent space have overlaps between different classes, these overlaps are considerably smaller than the vanilla VAE case. In the rVAE case the hearts suit overlapped with all the other three suits. Towards the center of the decoded latent space all four suits seem to have overlapped. Note that the manifolds corresponding to different classes cannot overlap (if objects are different), resulting in the characteristic "crossing" pattern.

At such high rotations and shears, the structural difference between classes is substantially small. The decoded latent space (Figure 6d) also shows a significant overlap between all four suits in this dataset leading to multiple ambiguous images in this plot. It also shows how image from one class smoothly emerge into an image from a different class. The decoded angle *vs.* the ground truth angle is show (Figure 6f) shows the presence of ambiguity in predicting the angle. The inherent symmetries in the datasets along with high shears makes images indecipherable. The ground truth shear plot shows a smooth variation of the image attribute for every class, even with high shear transformation. Even though the encoding of this dataset is superior to that of vanilla VAE, the classification of suits is still ambiguous in the latent space due to the overlapping. This is because the applied transformations completely deformed the images, and even the rVAE cannot classify the input images using only the unlabeled data. The complete analysis and plots of the results for all four cases can be found in 'rvae_cards.ipnyb' Jupyter notebook in the repository, accompanied by the manuscript.

**4.d. rVAE on experimental dataset**

To understand the applications rVAE on an experimental dataset, we have chosen the STEM image of silicon in graphene manipulated by e-beam activation.[50-54] The dataset consists of 50 snapshots of the graphene system during the e-beam manipulation. The graphene system has both crystalline and amorphous regions along with several point defects around the individual atoms. Figure 7a shows the both the carbon atoms and silicon atoms (bright dots) in zeroth snapshot of the dataset. As an initial step to obtain feature vectors for the rVAE analysis, we pass the entire dataset through a pre-trained deep convolutional neural network (DCNN) to obtain the atomic coordinates. The details and the workings of this DCNN are thoroughly discussed in these references[43,55]. The DCNN is a semantic segmentation network that classifies every pixel in the images as belonging to one of the three classes *viz.*, carbon atom, silicon atom, and the background. The output of the DCNN of the image shown in Figure 7b where the three classes are shown in three different colors. The atomic coordinates are then obtained by calculating the center of masses of the pixels of each atom. These atomic coordinates are shown in Figure 7c, where the carbon atoms are plotted as blue dots and the silicon atoms are plotted as bright red dots. These atomic coordinates are plotted on top of the original STEM image and as discussed earlier, the dataset of feature vectors for the identified atoms are constructed by cropping the image of the certain window size around these atoms. The window size is chosen such that it fits three hexagonal graphene rings around the central atom. It should be noted that the analysis that follows is sensitive to the window size used to create these feature vectors. The feature vectors are formed for all the identified atoms in all the snapshots which act as the input to the rVAE network. The rVAE network has 512 neurons in each of two hidden layers for both the encoder and decoder networks. We have used rotational and translational invariant version of VAE with two user selected latent dimensions ($z_1$ and $z_2$). This version of rVAE has a total of 5 latent dimensions *viz.*, one dimension for encoding angle ($\theta$), two dimensions for encoding translations in x and y directions ($\Delta x$ and $\Delta y$), and two user selected latent dimensions ($z_1$ and $z_2$) to encode the neighborhood of each atom in the dataset.

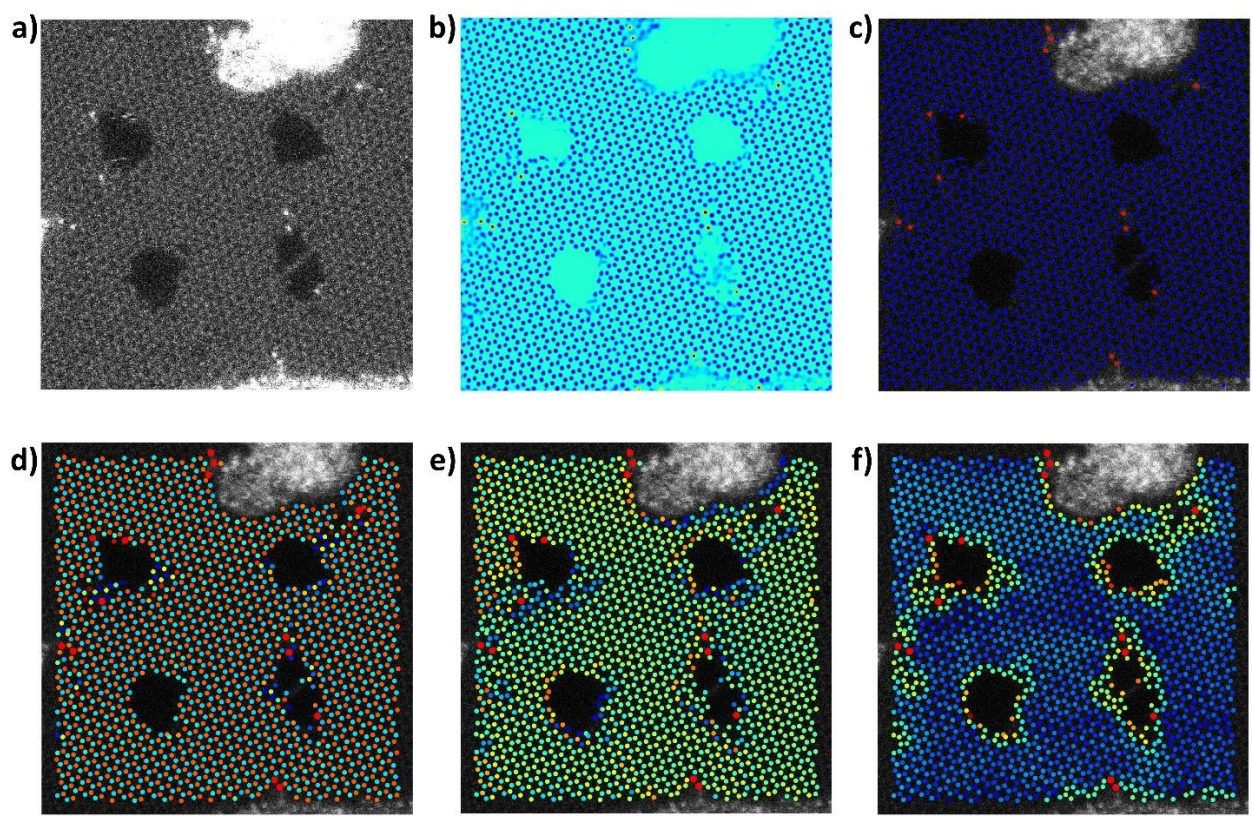

**Figure 7.** (a) Raw STEM image of the zeroth frame of the graphene dataset, (b) Semantically segmented output of the DCNN clearly discriminating two atom classes from the background class, (c) Atomic coordinates obtained from the center of masses of atom pixel clusters of the DCNN output. Each atom in the zeroth frame is colored using the encodings of (d) angle encoding ($\theta$), (e) first latent variable ($z_1$), and (f) second latent variables ($z_2$) of the rVAE network.

After the training phase of rVAE network, the zeroth frame is re-plotted in Figures 7d-f, where the atomic coordinates are now colored using the encodings of the feature vectors. Encoded angle, the first latent variable ($z_1$) and the second latent variable ($z_2$) are used for plotting the colors of the atomic coordinates in Figures 7d, 7e, and 7f respectively. For an ideal graphene layer, there are two symmetric unit cells around the central atom differed by 60°. The rVAE is able to identify the optimal rotational encodings for these two structures, as is represented by the checkerboard pattern in Figure 7d corresponding to the two rotational variants. The second latent variable ($z_2$) in in Figure 7f encodes the closeness to the amorphous region. It has a different value when the amorphous region is present in the cropped image of the atom which can be inferred by the sharp change in the color of this latent variable near the amorphous regions. First latent variable ($z_1$) seems to encode other forms of variability in the images like defects and strain in the feature vectors. The absolute values of the encodings do not contribute the understanding of the workings of the rVAE and hence the color bars for Figure 7d-f are omitted for brevity.

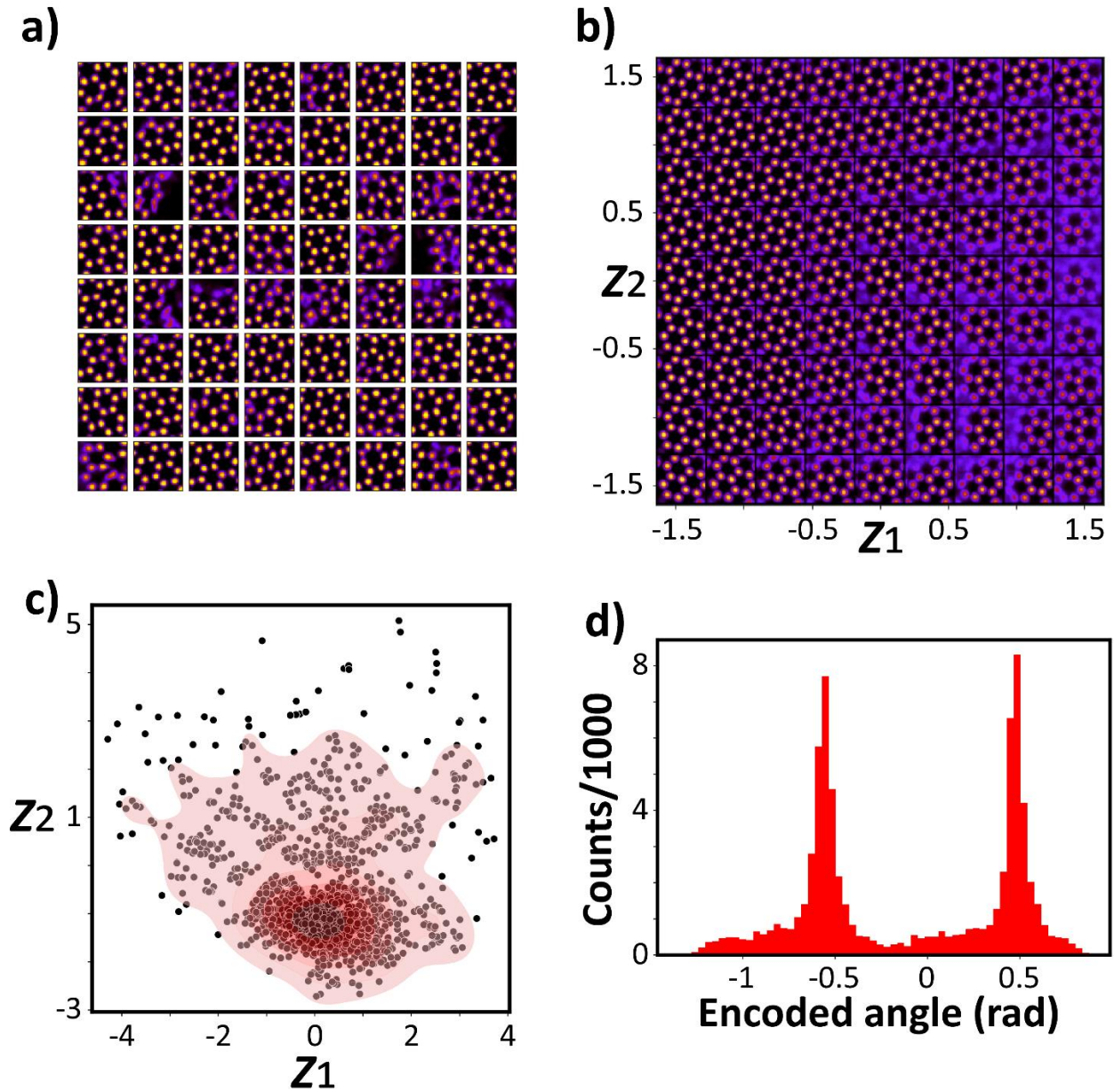

**Figure 8.** (a) sixty-four randomly picked images from the input to the rVAE network, (b) decoded latent space of the rVAE network trained on graphene data, (c) kernel density plot of the latent distributions, (d) histogram of angular encodings ($\theta$) in radians divided into 60 bins.

Further details of the rVAE encodings are discussed in Figure 8. Sixty-four randomly picked images (feature vectors) from the input dataset are shown in Figure 8a, which gives us insights about the training dataset. The decoded latent space for this network is shown in Figure 8b which elucidates the encodings as a function of the latent space. It can be seen from the decoded latent space that how certain areas encode the ideal graphene structures and the smooth variation into the images with amorphous regions. Although invisible to the naked eye, the other factor of variation in the images i.e., strain is also expected to be encoded smoothly in the latent space. Further analyses are required to show this which are beyond the scope of this article.

Figure 8c shows the kernel density plot of the latent distribution ($z_1$-$z_2$) of the atomic coordinates in the zeroth frame. The large cluster of points towards the center of the image corresponds to the ideal graphene structures as a majority of the input feature vectors correspond to them. The points away from the cluster are the feature vectors with defects and/or close to the amorphous regions. Finally, encoded angles for all the atoms found in the fifty frames are plotted as a histogram with 60 bins in Figure 8c. The two sharp peaks in the plot confirms that the two rotationally symmetric variants in the graphene layer are indeed found by the rVAE. These encodings in the 5 latent dimensions act as adequate feature vectors that encodes both the angular component and the neighborhood independently for each atom. The discussions on the latent dimensions corresponding to the translational components is omitted for brevity. The entire analysis can be visualized and reproduced using the Jupyter notebook 'rVAE_graphene.ipnyb' that accompanies the article.

## 5. Conditional VAEs

In this section, we discuss conditional rotationally invariant variational autoencoder, which we refer to as crVAE for the rest of the manuscript. In the conditional construction of VAEs, we assume that the input dataset is divided into discrete classes, and the class labels for each datapoint are known *a-priori*. Alternatively, condition can be continuous vector function. Generally, condition represents known factor of variability in the data.

First, we consider the discrete conditional rVAE. The decoder network, also known as the generator network, then learns to decode the encoded latent space conditioned on the class label of the data point. In the generator network, the encoded z-vector by rVAE is concatenated with the one hot encoded class label for the datapoint before decoding it with the spatial decoder network to form the crVAE. To make the encoder-decoder network symmetrical, we also concatenate the class label to the flattened input image. Since the decoder part is conditioned on the class labels, one latent space per class is learned independently by the crVAE. Since the construction of rVAE is already independently conditioned the rotation and translation encodings, for crVAE, the class labels can be sampled corresponding to one of the latent spaces, then independently sample a vector for the rotational and translational latent vector and random point in the latent space to generate images using the decoder net. It is a salient feature of the construction of crVAE that the latent representations can be conditioned on either discrete class labels or the continuous ground truth values that produced the dataset (shear in the case of cards dataset). The ELBO used for optimizing the crVAE is the same as the rVAE and is shown in the equation 4 and the schematic of the workings of crVAE are shown in Figure 5b.

**5.a. crVAE on the cards dataset**

To explore the machinery of the crVAE, we used the same four cards datasets discussed in the VAE section. For the architecture, we used 256 neurons in the two hidden layers of the encoder and the decoder for all four cases. The results for the first three cases are not discussed here as the explanations are straightforward and follow a similar trend to those of the fourth dataset. We chose the cards dataset iv as the rVAE in the previous section produced suboptimal encodings.

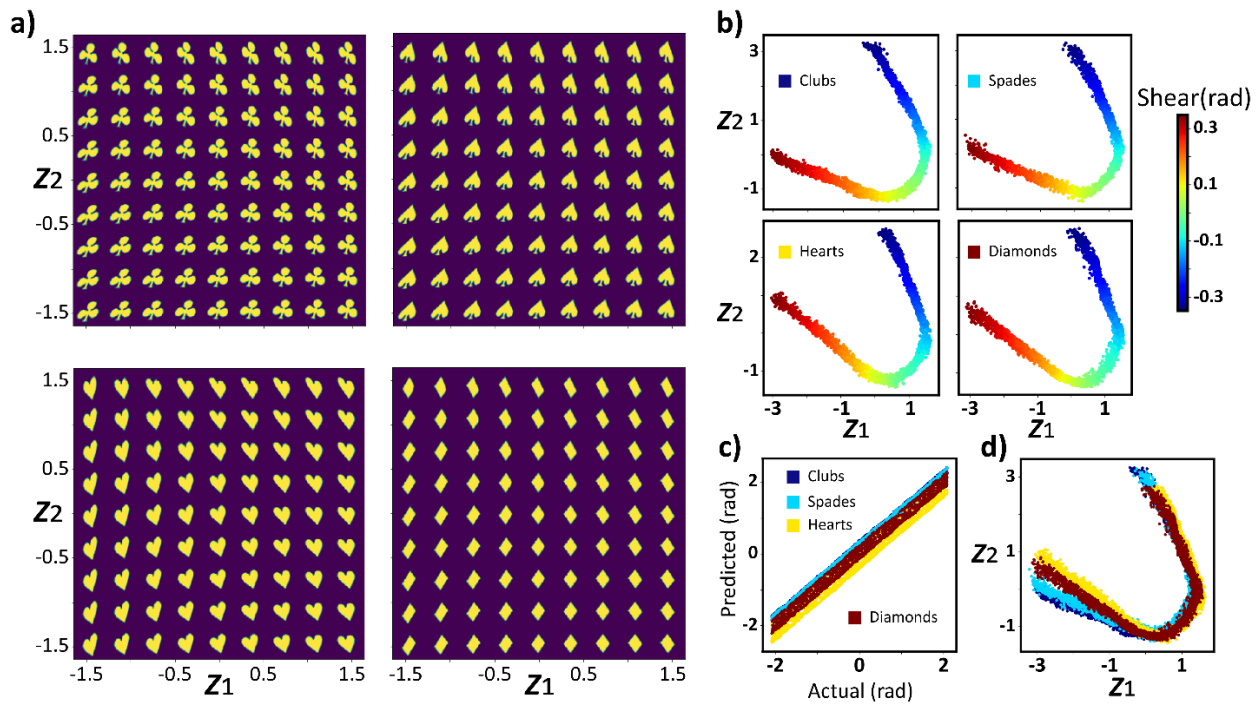

**Figure 9.** (a) Four independent latent spaces formed as a result of four classes in the input cards dataset-iv, (b) latent distributions of four classes colored with ground truth shear value, (c) encoded angle *vs.* ground truth angle used to rotate the input image, (d) latent distributions of four latent

spaces shown in a single plot and colored using the ground truth labels of each datapoint. The colorbar for (d) is same as the one shown in (c).

The crVAE analysis on the cards dataset iv are discussed in Figure 9. The decoded latent spaces for each class of clubs, spades, hearts, and diamonds are in Figure 9a. Since the class label is now not a feature of variablilty and the rotational component is already encoded by the rotational latent dimension $\theta$, the only factor of variability left in the dataset is the shear value applied on the image. This shear value is supposed to vary smoothly in the 2-dimensional latent space. This can be observed in latent distributions colored by the ground truth shear value in Figure 9b for all the four classes. The encoded angle plotted against the ground truth angle and is shown in Figure 9c. The encoded angle seems to be a single linear function of the ground truth angle with some variance which is due to the inherent symmetries present in the dataset. Finally, Figure 9d shows the latent distributions colored with class labels. Although these the encodings of different classes seem to overlap with each other, the encodings of each class live in a separate latent space conditioned on the class label. Hence, crVAE serves to be a better method of encoding the dataset of classes. However, this comes with a caveat of knowing the class labels *a priori*. The entire analysis of crVAE on all four cards datasets (i-iv) can be accessed through 'crvae_cards.ipnyb' Jupyter notebook that accompanies the manuscript.

**5.b. crVAE on the experimental dataset**

To elucidate the workings of crVAE on an experimental dataset, we used the graphene dataset discussed in the rVAE section. Since crVAE prerequisites the division of dataset into classes, we first divided the cropped images into classes based on the number and type of rings around the central atom. Every carbon atom in a graphene layer is a part of three hexagonal structures, only those three rings involving the central atom are used for classification. The classification task for this huge experimental dataset is time consuming and hence an automated workflow to classify individual sub-images is proposed. Initially, a graph of atoms identified by the DCNN network is formed. An edge is formed between two atoms if the distance between them is less than a pre-defined threshold value. This threshold value is set so that only the three nearest neighbors of each atom are connected. To accommodate for the defects in the graphene layer, this distance is selected to be more than the ideal C-C bond with a restriction that each atom can have a maximum of three layers. The graph for each snapshot is formed in one go and 5-7 membered rings are then identified in the whole image by finding cycles for a given graph. A dummy channel is created in which the central pixel of each identified ring is populated with a value equal to the ring size. This dummy channel is then added to the original STEM image before cropping the sub-images to form the input dataset. Each cropped sub-image is then classified using the information from the dummy channel. Although one can form a large number of classes based on the neighborhood of the central atom only six classes with the highest number of datapoints associated to them are considered for analysis. It should be noted that although this method of classifying the images is less time-consuming it can be error-prone while forming the edges in the graphs or due to improper identification of the atoms in the STEM image.

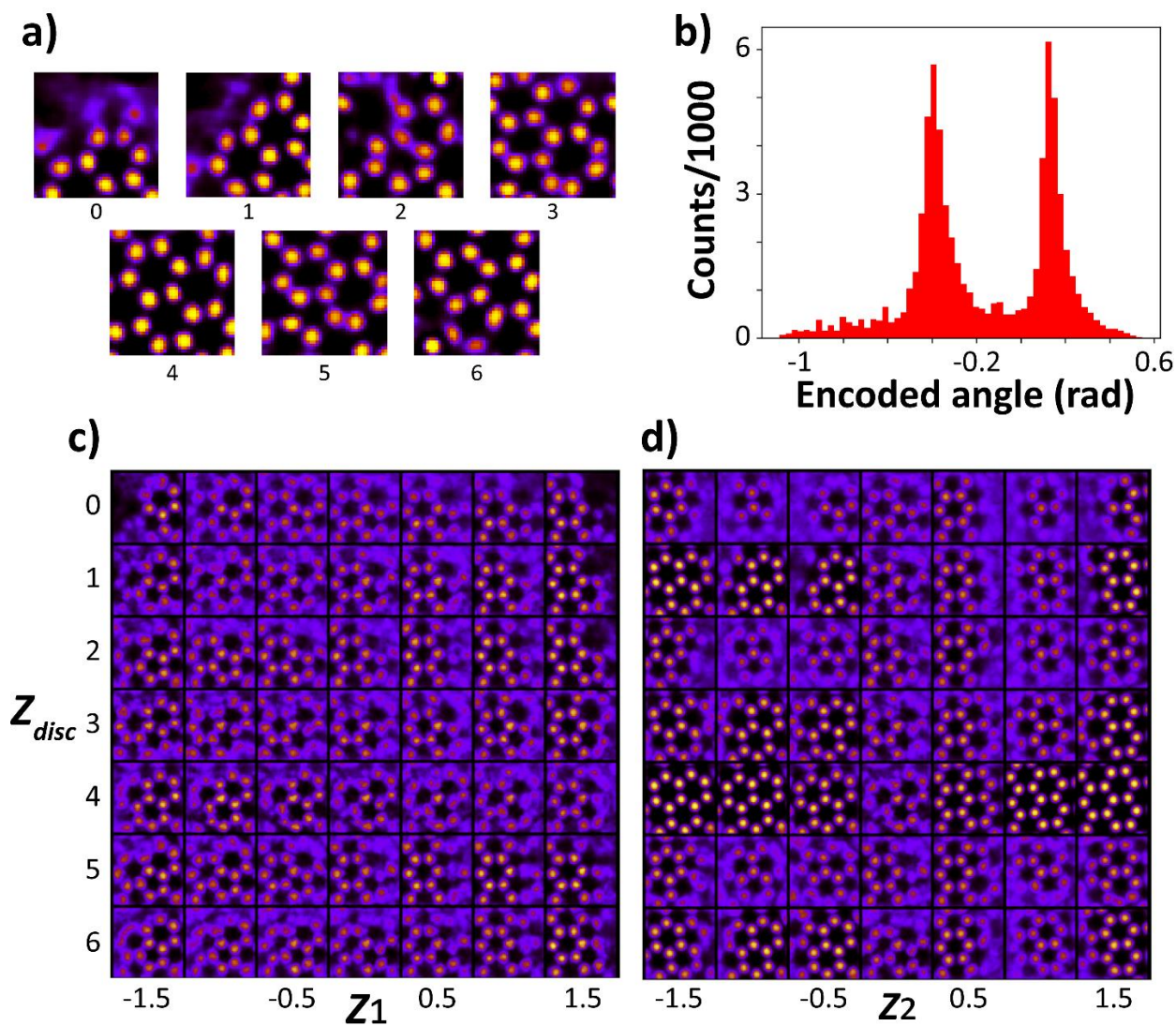

**Figure 10.** (a) Representative datapoints for all seven classes used in the crVAE network for graphene dataset, (b) histograms of angular encodings ($\theta$) divided into sixty bins, (c and d) traverse manifolds where one latent dimension along the *x*-axis and the class encodings along the *y*-axis are varied while the other latent variables are held constant at zero. First latent variable ($z_1$) is varied in (c), and the second latent variable ($z_2$) is varied in (d).

The seven classes considered for analysis are delineated in Figure 10a which are 0: images with fewer than 2 rings, 1: images with 2 [6, 6] rings, 2: images with 2 [5, 6] rings, 3: images with 3 [6, 6, 7] rings, 4: images with 3 [6, 6, 6] rings, 5: images with three [5, 6, 7] rings, and 6: images with 3 [5, 6, 6] rings. It should be noted that only the three hexagonal structures that are attached to the central atom are considered for classification and not the orientation of the rings. For example, in an image classified as class 6, the 5 membered ring can be present in any position. Firstly, a histogram of the encoded angles of the sub-images considered for this analysis is shown in Figure 10b. It has a characteristic twin peak similar to the case of rVAE which implies that the two rotational variants of the graphene sublayer are identified for all the classes in the dataset. Since this dataset with seven classes results in seven independent latent spaces, they are omitted for brevity. All seven manifolds can be visualized in the Jupyter notebook that accompanies this section.

For the concise presentation of latent spaces, a new way of visualizing the latent space for a dataset with discrete classes is used called the traverse manifold. In this method, all the continuous latent variables except for one are fixed and an image per class is shown along this dimension. The first of these traverse manifolds is shown in Figure 10c where $z_1$ is varied while the second one is shown in which $z_2$ is varied. The angular and translational latent dimensions are set to zero for both these manifolds. The latent variable which is being held constant is always set to zero. This method is similar to the decoded latent space of the rVAE where the rotational and translational encodings are set to zero. The disadvantage with this method of visualizing is that when no or few images from the dataset are encoded along the zero of the fixed dimensions, we see images that are not part of the training set in the traverse manifolds. For example, class 4 (row-5) in Figure 10d has the ideal graphene lattices but when $z_2$ is fixed at zero in Figure 10c, we see a lot of noise. This tells us that not much structural information is stored along $z_2 = 0$ for the class 4. But these traverse manifolds give us a condensed representation of how different classes are encoded in the latent dimensions.

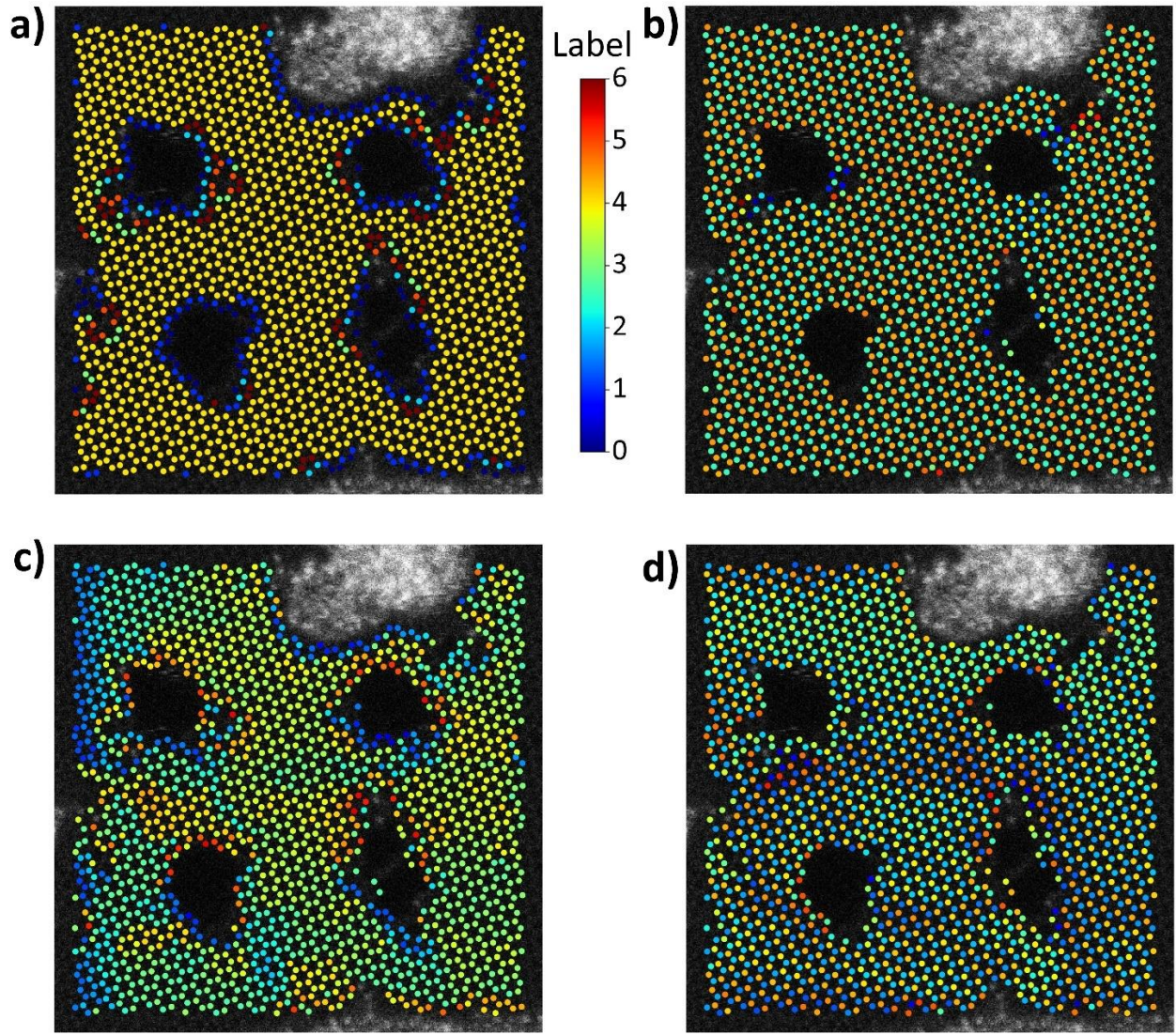

**Figure 11.** Atoms identified by the semantic segmentation network in the zeroth frame are colored using the (a) ground truth class labels, (b) angular encodings ($\theta$), (c) value of the first latent variable ($z_1$), and (d) values of the second latent variable ($z_2$) of the crVAE network trained on the graphene dataset.

Further results of the crVAE analysis on the graphene dataset are shown in Figure 11 in the real space. In Figure 11a, the atoms considered for the analysis in the zeroth frame of the dataset are colored with their associated class. These class labels act as the input to the crVAE. In Figure 11b, the atomic coordinates are colored with the encoded angle which shows a checkerboard type pattern line observed in rVAE. Finally, the values of the latent variables are used to color the atoms in Figure 11c ($z_1$) and 11d ($z_2$). Not much can be concluded from these encodings without knowing the variabilities in each class of the dataset. However, Figure 11d ($z_2$) also shows a checkboard type pattern which shows that some of the information from the angle in permeated into the second user selected latent variable $z_2$. This can happen when the $z_1$, $z_2$ latent variables encode the translational/rotational variabilities present in the dataset. For example, in this dataset, since the non-straining rotational and translational components are encoded by ($\theta$, $\Delta x$, and $\Delta y$), and the defects are classified into different classes, the only variability that is left in the images is the strain in the structures. If $z_1$ encodes this, there is not much information left for the other latent variable to encode. Decreasing the number of latent dimensions in such scenarios will lead to better interpretation of the latent variables. However, the encodings by crVAE will serve as better feature vectors that encode the neighborhood of each atom for any ML analysis than the corresponding ones of rVAE. This entire analysis of crVAE on the graphene dataset can be found at 'crvae_graphene.ipnyb' jupyter notebook that accompanies the manuscript.

## 6. Semi-supervised VAEs

Variational autoencoders are a class that belongs to the unsupervised family of machine learning algorithms. It is seldom the case that we encounter a training dataset whose data points are all divided into classes, and class labels for all the data points are known a priori. When they are known, the crVAE does an excellent job in projecting the dataset into a low dimensional latent space while explicitly disentangling the rotations and/or translations involved in the training data. The practically encountered scenario is where the data is available in the form of small, manually labeled datasets in published papers, catalogs, and other forms, and we aim to use these as a partially known labels. Alternatively, the small subset of data can be labeled manually by human, and then we aim to propagate this label across systems with a strong rotational or translational disorder. The ssrVAE (semi-supervised rotationally invariant autoencoder) generalizes from the small, labeled subset (supervised data) to the unsupervised data that is drawn from a different distribution of variations along with a strong rotational/translational disorder. To achieve this, the ssrVAE takes advantage of important features of rVAE and crVAE.

The architecture of ssrVAE has two encoders, namely *encoder_z* and *encoder_y*. When presented with the supervised dataset with labels, the *encoder_y* acts as a classifier network which is a multi-layer perceptron with *softmax* activation function where it predicts the class label for the datum, and the cross-entropy loss is used to train the *encoder_y* network. The *encoder_z* network is similar to the one we have encountered in the crVAE case, where it projects the image into low dimensional latent space while explicitly using 1-3 latent variables for disentangling the rotations and translations in the dataset. When presented with the unsupervised training data, the encoder_y tries and predicts the class label for each data point, and this class label is concatenated to the *z*-vector in the latent space, and the decoding is conditioned on the class label. The only difference between the unsupervised and supervised settings is that the class labels in supervised settings are used to train *encoder_y*, while class labels predicted by the *encoder_y* are used for the unsupervised case. Since ssrVAE tries to reconstruct the dataset using the class labels predicted by the *encoder_y* network, it takes advantage of the conditioning in crVAE while explicitly disentangling the rotational and/or translational latent vectors.

**6.a. ssrVAE on cards**

To delineate the workings of ssrVAE, we have used a small portion of the cards dataset-i (rotations = 12º, and shear = 1º) to try and predict the labels of the cards dataset-iii (rotations = 120º, and shear = 1º) and the cards dataset-iv (rotations = 120º, and shear = 20º). This way of choosing different distributions of datasets for supervised and unsupervised settings of ssrVAE gives us insights into how ssrVAE extrapolates the knowledge it learns from the supervised data to the unsupervised dataset. Supervised training set is constructed by choosing 200 images per card and applied transformations with the rotation and shear values randomly sampled from the cards dataset i. The unsupervised datasets comprise 3000 images per card with rotations and shear values from the cards dataset ii and iv. We trained two networks for these unsupervised datasets while holding the supervised dataset constant. The number of neurons used in the encoder_z and decoder layer for the high rotation and low shear case is 128, while for the high rotation and high shear case is 256 in each of the 2 hidden layers. Finally, to see how each network performed, two validation datasets (one for each case) with 1000 images per card suit are formed from the distributions that formed the unsupervised dataset. For brevity, the results for high rotations and high shear (cards dataset-iv) are only discussed below in this manuscript in Figure 12.

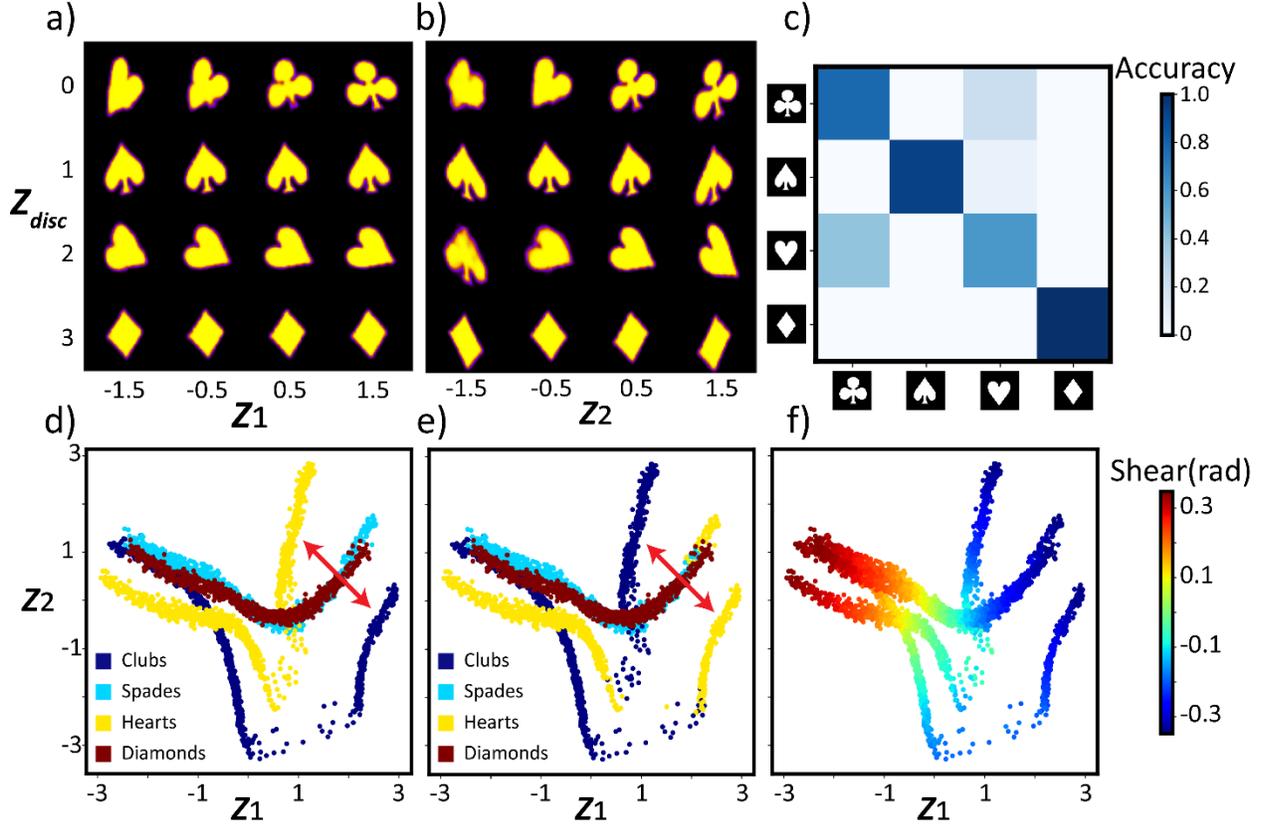

**Figure 12.** Results of ssrVAE trained on cards dataset where the unsupervised dataset is drawn from cards dataset-iv and supervised dataset is drawn from cards dataset-i. Traverse manifolds where (a) first latent variable is varied ($z_1$) and (b) second latent variable ($z_2$) is varied. (c) Confusion matrix on the validation dataset. Latent distributions of the validation colored with (d) ground truth labels, (e) labels predicted by the ssrVAE network, and (f) ground truth shear values of the data.

Figure 12a and 12b uses the traverse manifold to visualize the decoded latent spaces as discussed in the crVAE section. In Figure 12a, $z_2$ is held constant at zero while the other latent variable $z_1$ is varied along the *x*-axis and vice-versa in Figure 12b. The discrete class label is varied along the *y*-axis in both the traverse manifolds. To generate these manifolds, both encoder_y and encoder_z are discarded after the training phase and class labels of each class are concatenated to the sampled *z*-vector to decode the latent vectors. It can be observed that the class 0 (clubs, first row) and class 2 (hearts, third row) have some misclassifications. This can also be inferred from the normalized confusion matrix shown in Figure 12c. The accuracy rates of clubs and spades is around 50% with the images being identified as each other. The latent distributions of the validation dataset are plotted in Figures 12d and 12e and colored using the ground truth class labels in 12d and the predicted class labels in 12e.

The misclassification between clubs and spades are shown using red arrows in Figures 12d and 12e where the two branches in the latent distribution seem to have exchange colors. This is due to the large difference in distribution between the supervised and unsupervised datasets and the structural similarities between the highly sheared hearts and the clubs classes. Finally, the latent distributions of the validation dataset are colored using ground truth shear values in Figure 12f which shows a smooth variation of shear in the latent space. Even though the latent distributions of shear values of different classes seem to have overlapped, as discussed in the crVAE case, each class lives in an independent latent space. The latent space in which each of these encodings live is a function of the predicted class label by encoder_y which is denoted by the color in Figure 12e. It should be noted that in the supervised training dataset, the maximum value of shear present is 1º and the ssrVAE generalized it to the unsupervised dataset where the maximum value of shear is 20º. The trained ssrVAE has achieved and accuracy of 82% on the unsupervised dataset by generalizing the features learnt from the supervised dataset. In conclusion, ssrVAE, when presented with 800 images with low rotations (12º) and low shear (1º), is able to classify with 82% accuracy, explicitly disentangle the rotations, and form a smooth varying latent space of other variations of the high rotation (120º) and high shear (20º) dataset with 12,000 images. The complete analysis of the two datasets can be accessed and visualized using the 'ssrvae_cards.ipnyb' jupyter notebook that accompanies the manuscript.

**6.b. ssrVAE on graphene dataset**

Graphene dataset setting discussed in the crVAE section is used to illustrate the workings of ssrVAE as well. Only 15% of the input images that fall into one of the seven classes discussed in the crVAE section are used as the supervised dataset. A small portion (4%) of the dataset is set aside as the validation dataset. The remaining dataset including the images that do not fall into the seven categories discussed in the crVAE section are used as the unsupervised dataset. The traversal manifold way visualization is shown in Figure 13a and 13b. The first latent variable ($z_1$) is varied in Figure 13a while the $z_2$ is held constant at zero and vice-versa in Figure 13b. These manifolds provide insight into how different classes are encoded in the latent dimensions $z_1 - z_2$. The confusion matrix on the validation dataset is shown in Figure 13c. The network is able to predict the class-4 *i.e.*, the ideal [6, 6, 6] with the highest accuracy. The final accuracy on the validation dataset is 74.6%

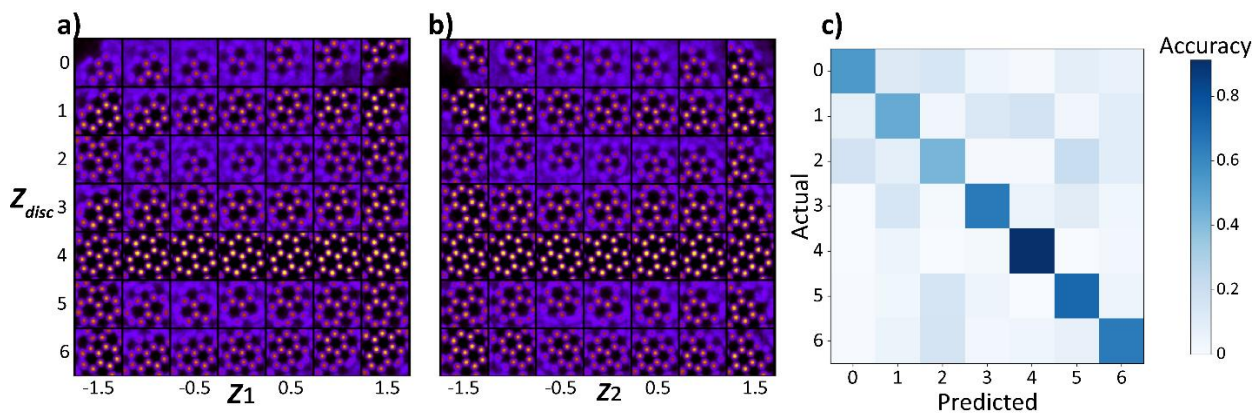

**Figure 13.** Traverse manifolds when (a) first latent variable ($z_1$) is varied and (b) second latent variable ($z_2$) is varied. (c) Confusion matrix of the validation dataset of graphene in the ssrVAE setting.

Further results of the latent encodings are shown in Figure 14. To obtain the encodings, the whole dataset is used as the input and each atom in the zeroth frame is colored using the encoded angle in Figure 14a. The checkerboard pattern observed in the previous sections is prevalent in this scenario as well. The latent dimensions $z_1$ and $z_2$ are used to color Figure 14b and 14c respectively. Since the ssrVAE is conditioned on the class predicted by *encoder_y*, these encoding values reside in different latent spaces. This means that the structural variability in the neighborhoods is already encoded by *encoder_y*. The latent dimensions as discussed in crVAE section encode the strain and microscopic variabilities which can be observed from the traversal manifolds in Figures 13a and 13b. Finally, the atoms are colored with the classes predicted by *encoder_y* in Figure 14d. A large portion of the frame is occupied by the ideal graphene lattice class 4.

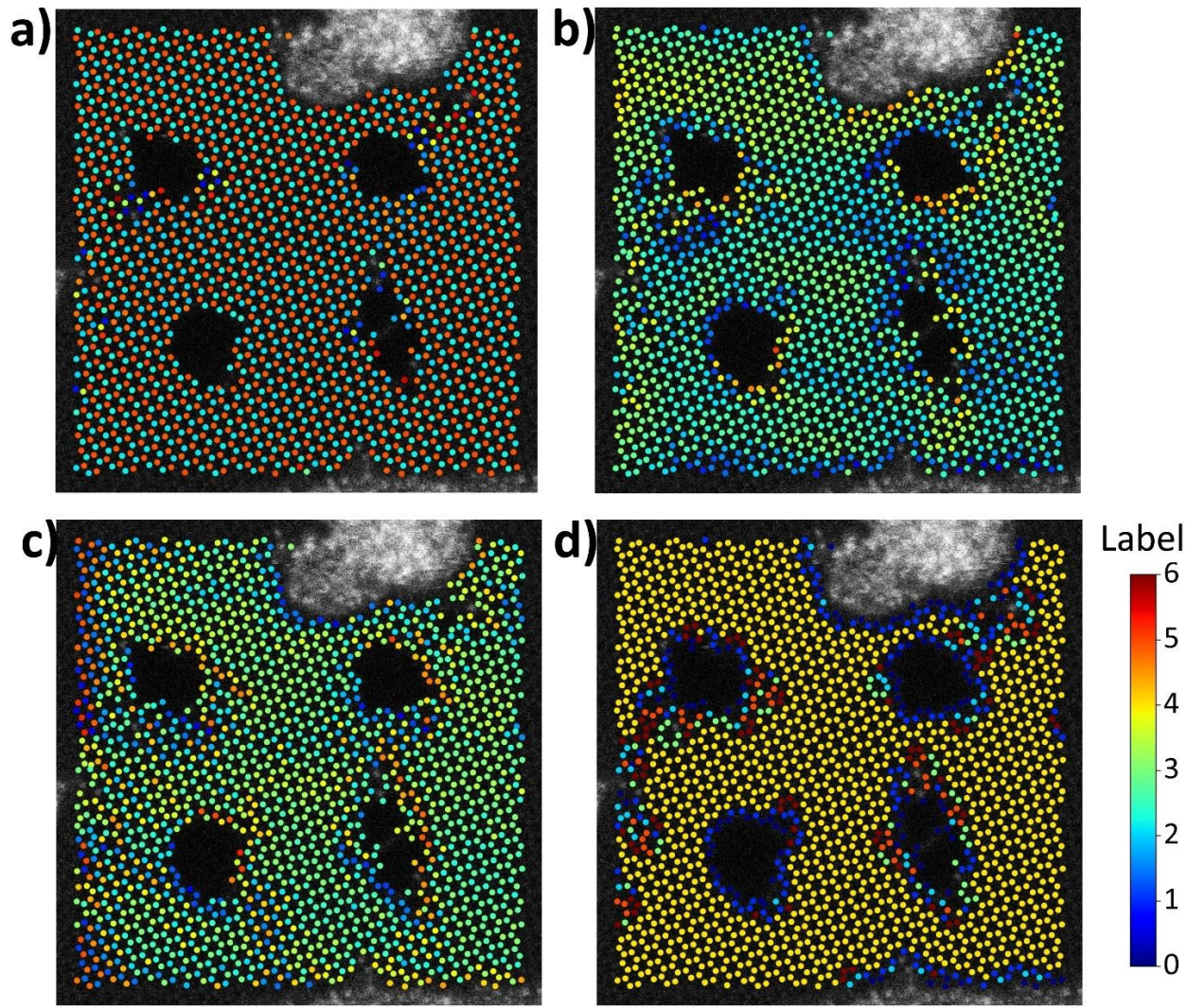

**Figure 14.** Atoms identified by the semantic segmentation network in the zeroth frame are colored using the (a) angular encodings ($\theta$), (b) value of the first latent variable ($z_1$), and (c) values of the second latent variable ($z_2$), (d) class labels predicted by the ssrVAE network trained on the graphene dataset.

## 7. Joint VAEs

Generally, the imaging data comprises objects from multiple classes with variability in images within the class. In supervised and semi-supervised machine learning settings, at least a part of the dataset is labeled. The algorithm then propagates the information learned from the known classes to the unlabeled dataset. However, the classes in the dataset are not apparent, especially in disciplines where a lot of labeled datasets are not readily available. Unsupervised clustering techniques divide the dataset into discrete classes based on the input dataset's hidden patterns or data groupings. The problem of learning low-dimensional manifolds while dividing an unlabeled dataset into discrete classes poses an ill-suited one for traditional machine algorithms.

This section discusses a class of invariant VAEs called the joint rotationally invariant variational autoencoders (jrVAEs) suited for solving this problem. The "joint" keyword in the name describes the aspect of the latent space of jrVAE where the latent space of the jrVAE jointly comprises discrete and continuous latent dimensions. This is similar to the unsupervised setting of the ssrVAE with the untrained classifier network. The discrete latent dimensions of the jrVAE are responsible for disentangling the dataset into various classes, while the continuous dimensions learn the in-class variations. These aspects can also be independently combined with the rotational and translational invariancy, where 1-3 latent dimensions capture the rotations and translations involved in the images.

The ELBO for the jrVAE is described by the equation 5 below, where reconstruction error refers to the error between the original and reconstructed image. The second term has three different components corresponding to the Kullback-Leibler (KL) divergence between the continuous latent variables and their corresponding priors. The first part $D_{KL}(q(z|x)||\mathcal{N}(0,I))$ is the KL divergence between the latent variables ($z$) and standard Normal prior distribution. This is followed by $D_{KL}(q(\theta|x)||M(0,\kappa))$ which is the KL divergence between the angle dimension ($\theta$) and the von-Mises prior with parameters 0 and $\kappa$. The prior distribution on the angle ($\theta$) dimension is a wrapped normal distribution which mimics a uniform distribution over $\theta$ in the interval [-π, π). The last term of the second term $D_{KL}(q(\Delta x|x)||\mathcal{N}(0,\sigma_s))$ refers to the KL divergence between the translational dimension and its normal prior with a standard deviation $\sigma_s$. Finally, the last term is the penalty on the discrete dimensions and is represented as the KL-divergence between the discrete latent variable and its prior $p(y) = Cat(y|\lambda)$ which is a discrete categorical distribution with parameters $\lambda$.

$$ELBO_{jrvae} = -Reconstruction\ error$$
$$- \gamma\big(D_{KL}(q(z|x)||\mathcal{N}(0,I)) + D_{KL}(q(\theta|x)||M(0,\kappa))$$
$$- D_{KL}(q(\Delta x|x)||\mathcal{N}(0,\sigma_s)) - C_z\big) - \gamma\big(D_{KL}(q(y|x)||p(y)) - C_y\big)$$

The hyperparameters of the ELBO that are to be optimized for a given dataset involve $\kappa$: parameter in the von-Mises prior on $\theta$, $\sigma_s$: standard deviation of the Normal prior on $\Delta x$, $\lambda$: parameters of the categorical prior distribution on the discrete latent variables. Along with these, the jrVAE network is highly sensitive to there are three other hyperparameters *viz.*, $\gamma$, $C_z$, and $C_y$. The continuous and discrete channel capacities are $C_z$ and $C_y$ respectively. When these are set to high values at a constant value of the weight coefficient $\gamma$, the effect of the KL terms diminishes and leads to a higher emphasis on the reconstruction error. The effect of these three terms is identical to that of parameter $\beta$ in the $\beta$-VAEs. During the initial steps of the training a low value of the channel capacities are used for a better entanglement of the data initially and then it is decreased for the reconstructions to improve. It should be thoroughly noted that although other variants of VAEs (mainly ssrVAE) are weakly dependent on the $\beta$ hyperparameters, the efficacy of jrVAEs, in our experience depends heavily on optimizing these parameters.

### 7.a. jrVAE on cards datasets

With the concise introduction to jrVAEs, in this section the cards datasets are used to delineate their capabilities. One jrVAE network for each of the four datasets are trained where 1024 neurons are used in hidden layers. A total of four hidden layers are used in both the encoder and decoder networks. Two user selected latent variables ($z_1$, $z_2$), three invariant latent variables ($\theta$, $\Delta x$, and $\Delta y$) and one discrete latent variable ($y$) with 4 dimensions comprise the latent space. The class label predicted by the jrVAE is the dimension of discrete latent variable with the highest value. The results of jrVAEs on cards datasets i and iii are discussed here. However, the results of all four datasets can be found in the Jupyter notebook that accompanies this section. The traverse manifolds for cards dataset i are shown in Figure 15a and 15b, where $z_1$ is varied in Figure 15a and $z_2$ is varied in Figure 15b while the rest of the continuous variables are held constant at zero. The traverse manifolds show that for the case of cards dataset i, jrVAE does an optimal job in identifying four classes without the a priori knowledge about classes. The four latent spaces, latent distribution, and angle plots can all be visualized in the Jupyter notebook. The traverse manifolds for cards dataset iii are shown in Figure 15c ($z_1$ is varied) and 15d ($z_2$ is varied). The jrVAE in this case does an ideal job for diamonds as the class corresponds to diamonds (class 3) is filled with all diamonds suit in a single orientation.

The network also identified all the clubs as a single class (class 0) albeit with multiple orientation. We have observed this trend of having multiple ground state zeros for the angular latent dimension in other variants of VAEs when the rotations in the dataset are high. Finally, the spades and the hearts dataset occupy both class 1 and class 3 due to the structural similarities present between them. We can also see that the hearts are identified with two different ground state orientations. Although, the classes predicted by the jrVAE arbitrary, each of the four classes is given a class label based on the majority ground truth classes. Using this predicted class labels, the accuracy of the jrVAE on the cards dataset iii is 82.11%. This calculation of accuracy is possible because we knew the ground truth class labels apriori. The complete analyses of all four datasets, latent spaces of all classes, latent distributions, angle plots, and shear variation in the latent space plots can be found in the 'jrVAE_card.ipnyb' jupyter notebook that accompanies the manuscript.

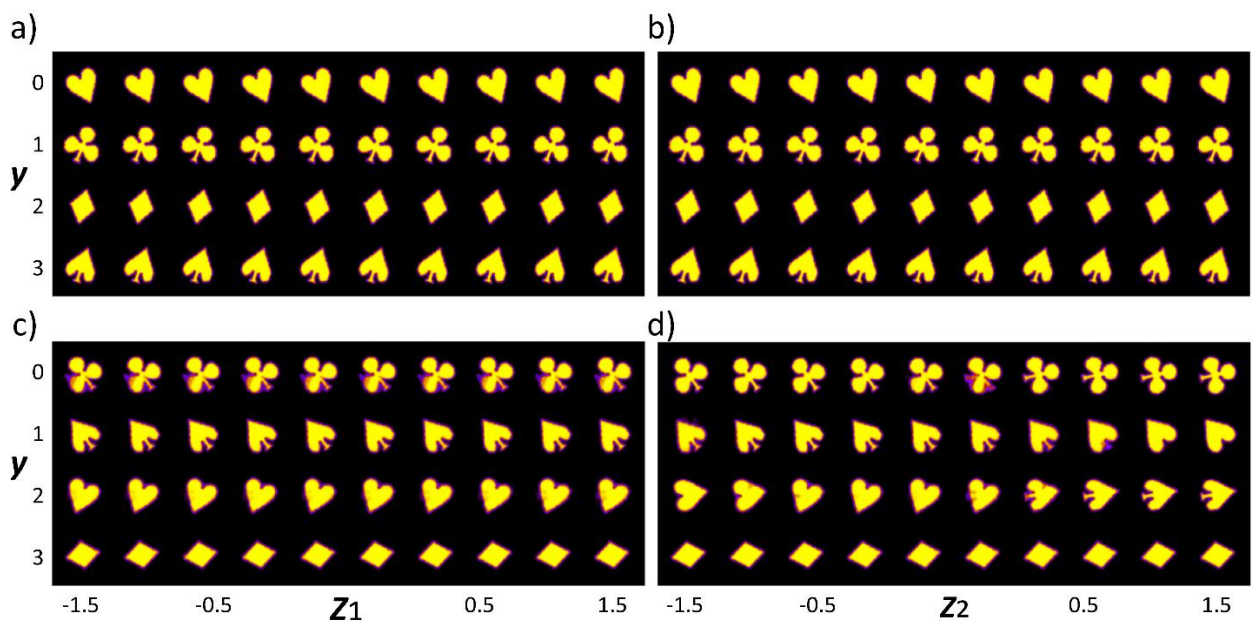

**Figure 15.** Traverse manifolds of the application of jrVAE on the cards dataset I when (a) $z_1$ is varied, (b) $z_2$ is varied, on cards dataset iii (c) $z_1$ is varied, (d) $z_2$ is varied.

As noted earlier, the jrVAEs are highly sensitive to hyperparameters $\gamma$, $C_z$, and $C_y$. These hyperparameters should be optimized for each dataset to produce ideal results for each dataset. In this review, we have used the default parameters that come with the AtomAI package used to create the jrVAE network. Optimizing these hyperparameters is beyond the scope of the review and an elegant way to optimize them for each dataset is proposed in ref.[56] The default parameters often result in sub-standard and can only be used as an initial step in understanding the dataset.

**7.b. jrVAE on graphene**

The results of the application of jrVAE on the graphene dataset are discussed in Figure 16. The network architecture used for this application is similar to that of the cards dataset with the only change in dimensionality of the discrete latent dimension ($y$) which is set to 2 for this case. The encodings in the angular dimension, $z_1$, and $z_2$ are shown on top of the detected atomic coordinates of zeroth frame in Figures 16a-c respectively. The checkerboard pattern in the angular latent dimension is also observed for jrVAE case as well which suggest that the jrVAE also found an optimal encoding for the angle. For the encodings of $z_1$ and $z_2$, since each subimage resides in either of the two independent latent spaces, these have to considered along with the class label to understand the variability along these dimensions. The predicted class label is used to color the atomic coordinates in Figure 16d, where large green dots correspond to the detected Si atoms. The traverse manifolds are shown in Figure 16e ($z_1$ is varied) and Figure 16f ($z_2$ is varied). Figures 16d-f show that the zeroth class (red dots in Figure 16d) corresponds to the ideal graphene lattice while the first class (blue dots in Figure 16e) corresponds to the defects and closeness to the amorphous regions. Conditioned on these classes, the latent variables $z_1$ and $z_2$ encode the additional variabilities in the feature vectors. The entire analysis can be reproduced and visualized in 'jrvae_graphene.ipnyb' jupyter notebook that accompanies the manuscript.

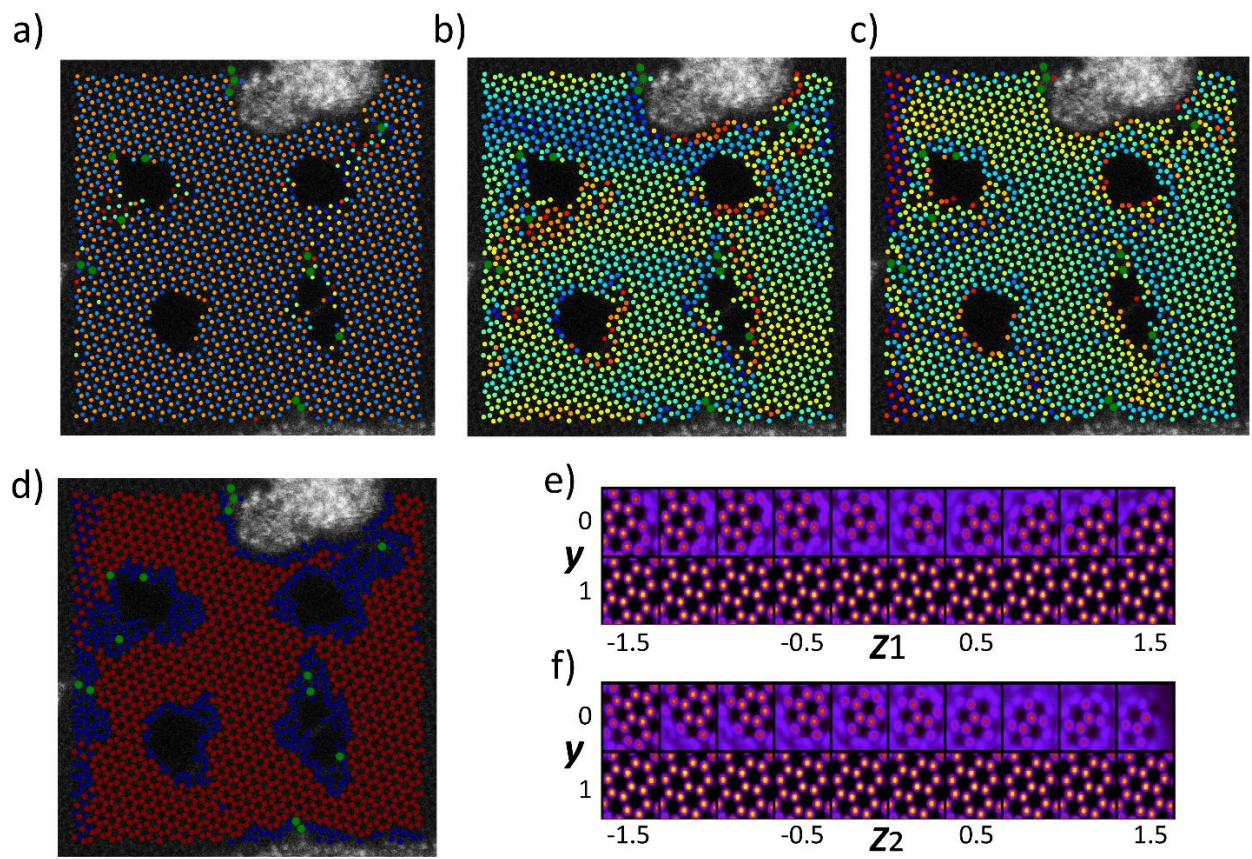

**Figure 16.** Atoms identified by the semantic segmentation network in the zeroth frame are colored using the (a) angular encodings ($\theta$), (b) value of the first latent variable ($z_1$), and (c) value of the second latent variable ($z_2$), and (d) class labels ($y$) predicted by the jrVAE network trained on the graphene dataset. Traverse manifolds of the jrVAE when (e) $z_1$ is varied and (f) $z_2$ is varied.

**Table I.** Autoencoders and Variational Autoencoders

| 1 | AE | Autoencoder, a type of neural network that learns to reconstruct input data by compressing it into a low-dimensional latent space and then expanding it back to the original dimensions. |
|---|---|---|
| 2 | Encoder | A neural network component in an autoencoder that maps input data into a lower-dimensional latent space representation. |
| 3 | Decoder | A neural network component in an autoencoder that reconstructs the original input data from the encoded representation. |

| 4 | Reconstruction loss | Measures the difference between the input and the reconstructed output data. It is typically computed using mean squared error or binary cross-entropy loss for image data and mean absolute error or cosine similarity for text data. |
|---|---|---|
| 5 | Latent distribution | Distribution formed by data when encoded into the latent space. Structure of the latent distribution allows to derive conclusions regarding physics of the system, number of classes, *etc*. |
| 6 | Decoded latent space | The image formed *via.*, decoding the latent vectors sampled over the grid of points and represented as a table of images. Latent representations illustrate how the data traits change across the latent space, hence providing interpretability for latent variables. |
| 7 | VAE | Variational autoencoders, Bayesian generative models that leverage a probabilistic encoder and decoder to learn a latent representation of the data distribution and enable the estimation of posterior probabilities over latent variables. |
| 8 | Encoder-decoder network | The neural network architecture where encoded and decoded object have different dimensionality. The variational encoders-decoders can be used for building structure-property relationships.[38,57,58] |
| 8 | Prior distribution | Prior distribution of the latent variables in VAE. Usually assumed to be Gaussian. |
| 9 | KL divergence | Kullback-Leibler divergence, a measure of the difference between two probability distributions, commonly used in variational inference to measure the distance between a target distribution and an approximation. |
| 10 | Factors of variability | Traits in data. For example, in MNIST it is handwriting style and in cards it's the rotations and shears applied to the images |
| 11 | Disentanglement of representations | The capability of VAE to discover factors of variability in data. While lacking the rigorous definition, this is one of the strongest advantages of VAE and is immediately visible from data. Ideally, representations will discover factors of variability in data. |

| 12 | iVAE | Invariant-VAE, separates the physical factors of variability such as rotations, translations, or expansion/contraction in dedicated latent variables. |
| --- | --- | --- |
| 13 | rVAE | Rotationally invariant-VAE, a class of iVAE that explicitly disentangles rotations and/or translations in dedicated latent variables. Typically used for image data. |
| 14 | cVAE | conditional VAE, known physical behaviors such as object classes or continuous variables can be used to condition the data. For discrete labels, we form independent latent spaces for each class. |
| 15 | ssVAE | semi supervised VAE, works on the dataset when only partial targets/class labels are available. ssVAE propagates the information learnt from the supervised dataset to unsupervised dataset. Invariant ssVAE are extremely useful for building libraries of the images like pollen, stars, molecules. |
| 16 | jVAE | joint VAE, simultaneously separates objects into sub-spaces (classification) and disentangles representations within each space. |
| 17 | Traversal representation | Decoded latent space representation of joint VAE or ssVAE as a function of discrete class and one of the continuous latent variables when the rest of the latent variables are held constants at zero. |
| 18 | Channel capacity | Weights assigned to the channels of ELBO at each iteration during training. These capacities determine the term that dominates the ELBO during the training phase. |

## 8. Summary

To summarize, here we discuss the basic principles, implementations, and applications of the invariant variational autoencoders for the imaging data. The full list of introduced definitions is summarized in Table I. The central concept of VAE is the disentanglement of the representations of the data, *i.e.*, finding the parsimonious factors of variability. In the classical data sets like MNIST, the factors of variability refer to the characteristics of handwriting style such as tilt and width of letters. In imaging data, factors of variability include shape variations of nanoparticles, distortions of the unit cells due to strain or ferroelectric polarization, chemical neighborhoods of atoms in graphene, or variability of the local diffraction patterns in 4D STEM. Correspondingly, VAEs offer a potential pathway to learn these behaviors from observational data.

However, the direct applications of the simple VAEs to the experimental data is limited. First and foremost, in observational studies multiple objects of the same type can be observed at different observation angles. Second, the sampling procedure to identify the sub-images can be associated with the random shifts. Finally, the system can be associated with the presence of affine distortions, that can stem from intrinsic material structure, or be due to the imaging system. The shift- and rotation invariant VAEs introduced here allow separating these physical factors of variability directly and disentangle the remaining factors of variability in latent representations.

Similarly, very often the imaging data is collected from system with multiple discrete object classes, that can either be known, partially known (have examples), or unknown. We discuss conditional, semi-supervised, and joint VAEs as frameworks to address these tasks. Notably, conditional VAEs can also be extended towards the continuous conditional vectors, allowing to incorporate known physical factors of variability.

Jointly, this framework allows comprehensive analysis of imaging data sets. We further discuss the extensions of the VAE approaches towards representation learning, introducing physics-based loss functions, and causal representations. Some of the intuition beyond the formation and properties of the latent spaces and latent manifolds is discussed. Finally, we note that VAE approach can be extended to more complex tasks of building structure property relations (refs) and active learning but defer these to future publications.

The notebooks illustrating the concepts in this review are fully available https://github.com/saimani5/VAE-tutorials. We encourage the readers to explore these concepts further and apply VAE to own data.

**Acknowledgements:**
This work (workflow development, manuscript writing) was supported by the US Department of Energy, Office of Science, Office of Basic Energy Sciences, as part of the Energy Frontier

Research Centers program: CSSAS—The Center for the Science of Synthesis Across Scales—under Award No.DE-SC0019288, located at University of Washington, DC. This effort was also partially supported as part of the center for 3D Ferroelectric Microelectronics (3DFeM), an Energy Frontier Research Center funded by the U.S. Department of Energy (DOE), Office of Science, Basic Energy Sciences under Award Number DE-SC0021118. The authors thank Dr. Maxim Ziatdinov for authoring the python packages atomai and pyroved used in constructing the variational autoencoder networks discussed in the article. We also thank him for insightful discussions throughout this research.

# References

1  van Zuylen, J. The microscopes of Antoni van Leeuwenhoek. *Journal of microscopy* **121**, 309-328 (1981).
2  Masters, B. R. History of the optical microscope in cell biology and medicine. *eLS* (2008).
3  Fan, Z. *et al.* In situ transmission electron microscopy for energy materials and devices. *Advanced Materials* **31**, 1900608 (2019).
4  Bruma, A. *Scanning Transmission Electron Microscopy: Advanced Characterization Methods for Materials Science Applications*. (CRC Press, 2020).
5  Wang, Y., Skaanvik, S. A., Xiong, X., Wang, S. & Dong, M. Scanning probe microscopy for electrocatalysis. *Matter* **4**, 3483-3514 (2021).
6  Wen, H., Cherukara, M. J. & Holt, M. V. Time-resolved X-ray microscopy for materials science. *Annual Review of Materials Research* **49**, 389-415 (2019).
7  Hui, F. & Lanza, M. Scanning probe microscopy for advanced nanoelectronics. *Nature electronics* **2**, 221-229 (2019).
8  Gerber, C. & Lang, H. P. How the doors to the nanoworld were opened. *Nature Nanotechnology* **1**, 3-5, doi:10.1038/nnano.2006.70 (2006).
9  Binnig, G. & Rohrer, H. SCANNING TUNNELING MICROSCOPY. *Helvetica Physica Acta* **55**, 726-735 (1982).
10  Stroscio, J. A., Feenstra, R. M. & Fein, A. P. ELECTRONIC-STRUCTURE OF THE SI(111)2X1 SURFACE BY SCANNING-TUNNELING MICROSCOPY. *Physical Review Letters* **57**, 2579-2582, doi:10.1103/PhysRevLett.57.2579 (1986).
11  Asenjo, A., Gomezrodriguez, J. M. & Baro, A. M. CURRENT IMAGING TUNNELING SPECTROSCOPY OF METALLIC DEPOSITS ON SILICON. *Ultramicroscopy* **42**, 933-939, doi:10.1016/0304-3991(92)90381-s (1992).
12  Grutter, P., Liu, Y., LeBlanc, P. & Durig, U. Magnetic dissipation force microscopy. *Applied Physics Letters* **71**, 279-281 (1997).
13  Martin, Y. & Wickramasinghe, H. K. MAGNETIC IMAGING BY FORCE MICROSCOPY WITH 1000-A RESOLUTION. *Applied Physics Letters* **50**, 1455-1457, doi:10.1063/1.97800 (1987).
14  Noy, A., Vezenov, D. V. & Lieber, C. M. Chemical force microscopy. *Annual Review of Materials Science* **27**, 381-421, doi:10.1146/annurev.matsci.27.1.381 (1997).
15  Gruverman, A., Auciello, O. & Tokumoto, H. Imaging and control of domain structures in ferroelectric thin films via scanning force microscopy. *Annual Review of Materials Science* **28**, 101-123, doi:10.1146/annurev.matsci.28.1.101 (1998).
16  Kim, M. *et al.* Nonstoichiometry and the electrical activity of grain boundaries in SrTiO3. *Physical Review Letters* **86**, 4056-4059, doi:10.1103/PhysRevLett.86.4056 (2001).
17  Pennycook, S. J., Varela, M., Lupini, A. R., Oxley, M. P. & Chisholm, M. F. Atomic-resolution spectroscopic imaging: past, present and future. *Journal of Electron Microscopy* **58**, 87-97, doi:10.1093/jmicro/dfn030 (2009).
18  Sohlberg, K., Rashkeev, S., Borisevich, A. Y., Pennycook, S. J. & Pantelides, S. T. Origin of anomalous Pt-Pt distances in the Pt/alumina catalytic system. *Chemphyschem* **5**, 1893-1897, doi:10.1002/cphc.200400212 (2004).
19  Chisholm, M. F., Luo, W. D., Oxley, M. P., Pantelides, S. T. & Lee, H. N. Atomic-Scale Compensation Phenomena at Polar Interfaces. *Physical Review Letters* **105**, doi:197602 10.1103/PhysRevLett.105.197602 (2010).
20  Browning, N. D. *et al.* The influence of atomic structure on the formation of electrical barriers at grain boundaries in SrTiO3. *Applied Physics Letters* **74**, 2638-2640, doi:10.1063/1.123922 (1999).

21    Clausen-Schaumann, H., Seitz, M., Krautbauer, R. & Gaub, H. E. Force spectroscopy with single bio-molecules. *Current Opinion in Chemical Biology* **4**, 524-530, doi:10.1016/s1367-5931(00)00126-5 (2000).
22    Rief, M., Oesterhelt, F., Heymann, B. & Gaub, H. E. Single molecule force spectroscopy on polysaccharides by atomic force microscopy. *Science* **275**, 1295-1297, doi:10.1126/science.275.5304.1295 (1997).
23    Balke, N., Bdikin, I., Kalinin, S. V. & Kholkin, A. L. Electromechanical Imaging and Spectroscopy of Ferroelectric and Piezoelectric Materials: State of the Art and Prospects for the Future. *Journal of the American Ceramic Society* **92**, 1629-1647, doi:10.1111/j.1551-2916.2009.03240.x (2009).
24    Van der Walt, S. *et al.* scikit-image: image processing in Python. *PeerJ* **2**, e453 (2014).
25    Krizhevsky, A., Sutskever, I. & Hinton, G. E. Imagenet classification with deep convolutional neural networks. *Communications of the ACM* **60**, 84-90 (2017).
26    Opitz, D. & Maclin, R. Popular ensemble methods: An empirical study. *Journal of artificial intelligence research* **11**, 169-198 (1999).
27    Vyas, A. *et al.* in *Proceedings of the European Conference on Computer Vision (ECCV).*  550-564.
28    Xia, G. & Bouganis, C.-S. On the Usefulness of Deep Ensemble Diversity for Out-of-Distribution Detection. *arXiv preprint arXiv:2207.07517* (2022).
29    Liu, Y., Kelley, K. P., Funakubo, H., Kalinin, S. V. & Ziatdinov, M. Exploring physics of ferroelectric domain walls in real time: deep learning enabled scanning probe microscopy. *Advanced Science* **9**, 2203957 (2022).
30    Liu, Y. *et al.* Disentangling electronic transport and hysteresis at individual grain boundaries in hybrid perovskites via automated scanning probe microscopy. *arXiv preprint arXiv:2210.14138* (2022).
31    Kingma, D. P. & Welling, M. Auto-encoding variational bayes. *arXiv preprint arXiv:1312.6114* (2013).
32    Kingma, D. P. & Welling, M. An introduction to variational autoencoders. *Foundations and Trends® in Machine Learning* **12**, 307-392 (2019).
33    Rezende, D. J., Mohamed, S. & Wierstra, D. in *International conference on machine learning.*  1278-1286 (PMLR).
34    Taud, H. & Mas, J. Multilayer perceptron (MLP). *Geomatic approaches for modeling land change scenarios*, 451-455 (2018).
35    Goodfellow, I., Bengio, Y. & Courville, A. in *Deep learning* Vol. 2016    330-372 (MIT Press Cambridge, MA, USA, 2016).
36    Vaswani, A. *et al.* Attention is all you need. *Advances in neural information processing systems* **30** (2017).
37    de Haan, P., Cohen, T. S. & Welling, M. Natural graph networks. *Advances in neural information processing systems* **33**, 3636-3646 (2020).
38    Kalinin, S. V., Kelley, K., Vasudevan, R. K. & Ziatdinov, M. Toward decoding the relationship between domain structure and functionality in ferroelectrics via hidden latent variables. *ACS Applied Materials & Interfaces* **13**, 1693-1703 (2021).
39    Lecun, Y., Bottou, L., Bengio, Y. & Haffner, P. Gradient-based learning applied to document recognition. *Proceedings of the IEEE* **86**, 2278-2324, doi:10.1109/5.726791 (1998).
40    Duvenaud, D. K. *et al.* Convolutional networks on graphs for learning molecular fingerprints. *Advances in neural information processing systems* **28** (2015).
41    Weininger, D. SMILES, a chemical language and information system. 1. Introduction to methodology and encoding rules. *Journal of Chemical Information and Computer Sciences* **28**, 31-36, doi:10.1021/ci00057a005 (1988).

42	Krenn, M., Häse, F., Nigam, A., Friederich, P. & Aspuru-Guzik, A. Self-referencing embedded strings (SELFIES): A 100% robust molecular string representation. *Machine Learning: Science and Technology* **1**, 045024, doi:10.1088/2632-2153/aba947 (2020).
43	Liu, Y., Proksch, R., Wong, C. Y., Ziatdinov, M. & Kalinin, S. V. Disentangling Ferroelectric Wall Dynamics and Identification of Pinning Mechanisms via Deep Learning. *Advanced Materials* **33**, 2103680, doi:https://doi.org/10.1002/adma.202103680 (2021).
44	Kalinin, S. V., Steffes, J. J., Liu, Y., Huey, B. D. & Ziatdinov, M. Disentangling ferroelectric domain wall geometries and pathways in dynamic piezoresponse force microscopy via unsupervised machine learning. *Nanotechnology* **33**, 055707 (2021).
45	Liu, Y., Ziatdinov, M. & Kalinin, S. V. Exploring causal physical mechanisms via non-gaussian linear models and deep kernel learning: applications for ferroelectric domain structures. *ACS nano* **16**, 1250-1259 (2021).
46	Valleti, S. M. P., Ignatans, R., Kalinin, S. V. & Tileli, V. Decoding the Mechanisms of Phase Transitions from In Situ Microscopy Observations. *Small* **18**, 2104318, doi:https://doi.org/10.1002/smll.202104318 (2022).
47	Liu, Y. *et al.* Decoding the shift-invariant data: applications for band-excitation scanning probe microscopy. *Machine Learning: Science and Technology* **2**, 045028 (2021).
48	Liu, Y., Huey, B. D., Ziatdinov, M. A. & Kalinin, S. V. Physical discovery in representation learning via conditioning on prior knowledge: applications for ferroelectric domain dynamics. *arXiv preprint arXiv:2203.03122* (2022).
49	Bepler, T., Zhong, E., Kelley, K., Brignole, E. & Berger, B. Explicitly disentangling image content from translation and rotation with spatial-VAE. *Advances in Neural Information Processing Systems* **32** (2019).
50	Kalinin, S. V., Dyck, O., Jesse, S. & Ziatdinov, M. Exploring order parameters and dynamic processes in disordered systems via variational autoencoders. *Science Advances* **7**, eabd5084, doi:doi:10.1126/sciadv.abd5084 (2021).
51	Ziatdinov, M. *et al.* Building and exploring libraries of atomic defects in graphene: Scanning transmission electron and scanning tunneling microscopy study. *Science Advances* **5**, eaaw8989, doi:doi:10.1126/sciadv.aaw8989 (2019).
52	Dyck, O., Kim, S., Kalinin, S. V. & Jesse, S. Placing single atoms in graphene with a scanning transmission electron microscope. *Applied Physics Letters* **111**, 113104, doi:10.1063/1.4998599 (2017).
53	Dyck, O., Kim, S., Kalinin, S. V. & Jesse, S. E-beam manipulation of Si atoms on graphene edges with an aberration-corrected scanning transmission electron microscope. *Nano Research* **11**, 6217-6226, doi:10.1007/s12274-018-2141-6 (2018).
54	Dyck, O. *et al.* Building Structures Atom by Atom via Electron Beam Manipulation. *Small* **14**, 1801771, doi:https://doi.org/10.1002/smll.201801771 (2018).
55	Ziatdinov, M. *et al.* Deep learning of atomically resolved scanning transmission electron microscopy images: chemical identification and tracking local transformations. *ACS nano* **11**, 12742-12752 (2017).
56	Biswas, A., Vasudevan, R., Ziatdinov, M. & Kalinin, S. V. Optimizing training trajectories in variational autoencoders via latent Bayesian optimization approach*. *Machine Learning: Science and Technology* **4**, 015011, doi:10.1088/2632-2153/acb316 (2023).
57	Roccapriore, K. M., Ziatdinov, M., Cho, S. H., Hachtel, J. A. & Kalinin, S. V. Predictability of Localized Plasmonic Responses in Nanoparticle Assemblies. *Small* **17**, 2100181, doi:https://doi.org/10.1002/smll.202100181 (2021).

58	Ziatdinov, M., Ghosh, A., Wong, C. Y. & Kalinin, S. V. AtomAI framework for deep learning analysis of image and spectroscopy data in electron and scanning probe microscopy. *Nature Machine Intelligence* **4**, 1101-1112, doi:10.1038/s42256-022-00555-8 (2022).